\documentclass[a4paper,11pt]{article}

\usepackage[top=13mm, bottom=13mm, left=13mm, right=13mm, footskip=7mm]{geometry} % Set page margins and footer spacing
\usepackage[T1]{fontenc}
\usepackage{lmodern}      % Latin Modern font family
\usepackage{indentfirst} % Indent the first paragraph of each section
\usepackage{lineno}
\usepackage[none]{hyphenat}
\usepackage{afterpage} % allows the current page of text to fill up and finish normally
\usepackage{placeins} % ensure that your tables never float into a new section, 
\usepackage{amsmath,amsfonts,amssymb,amsthm,mathtools} % Advanced math environments and symbols
\usepackage{upquote}     % Ensures correct rendering of backticks in code/verbatim

\usepackage{authblk}     % Enhanced author and affiliation block
\usepackage{graphicx}    % Core graphics support
\graphicspath{{figures/}}% Default folder for compiling images
\usepackage{tikz}
\usetikzlibrary{arrows.meta,backgrounds,calc,positioning,shapes.geometric}
\usepackage{caption}     % Advanced caption formatting
\usepackage{array,tabularx,tabulary,booktabs, pdflscape, ragged2e, rotating} 

\usepackage{cite}        % Standardized citation handling

\let\OLDthebibliography\thebibliography
\renewcommand\thebibliography[1]{
  \OLDthebibliography{#1}
  \setlength{\parskip}{0pt} % Removes space between paragraphs within an item
  \setlength{\itemsep}{0pt plus 0.3ex} % Reduces space between separate reference items
}

\usepackage{xcolor}      % Color definitions
\usepackage{fontawesome5}% Icons (e.g., GitHub logo)
\usepackage{hyperref}    % Hyperlinks (must generally be loaded last)
\usepackage{orcidlink}   % ORCID icons and links

\definecolor{VividBlue}{RGB}{0,102,204}
\definecolor{AcademicNavy}{RGB}{0,50,95}
\hypersetup{
  colorlinks=true,
  linkcolor=VividBlue,
  urlcolor=VividBlue,
  citecolor=AcademicNavy
}

\newcommand{\doilink}[1]{%
  \href{https://doi.org/#1}{\nolinkurl{doi:#1}}%
}

\definecolor{zenodoblue}{HTML}{016292}

\definecolor{githubgray}{HTML}{333333}

\definecolor{arxivred}{HTML}{B31B1B}

\begin{document}

% ==============================================================================
% TITLE AND AUTHORS
% ==============================================================================
\title{\LARGE \textbf{Wind Turbine Maintenance Log Labelling Framework}: \\ \Large LLM-Driven Data Correction and Enrichment \\ via Semantic Extraction of Reliability Intelligence}

\author[1]{Max Malyi\orcidlink{0000-0002-1503-9798}\thanks{Corresponding author: \url{Max.Malyi@ed.ac.uk}}}
\author[1]{Jonathan Shek\orcidlink{0000-0001-5734-2907}}
\author[1]{Alasdair McDonald\orcidlink{0000-0002-2238-3589}}
\author[2]{André Biscaya\orcidlink{0000-0002-8158-4284}}

\affil[1]{Institute for Energy Systems, School of Engineering, The University of Edinburgh, Edinburgh, UK}
\affil[2]{Nadara, Lisbon, Portugal}

\date{November 2026}

\maketitle

% ==============================================================================
% ABSTRACT
% ==============================================================================
\begin{abstract}
\phantomsection
\label{sec:abstract}
\noindent Inconsistent codes and unstructured text limit the use of wind turbine maintenance records in reliability analysis. This study presents a topology-aware large language model (LLM) workflow to correct, enrich and structure these records. The workflow processed 16,316~records from 280~turbines across 31~onshore wind farms over 9.2~years. Maintenance narratives and equipment context guide system-code correction and maintenance-type relabelling. System- and topology-specific record batches yield candidate action and failure-mode taxonomies for record-level labelling. Failure-mode profiles link labels to common mechanisms, possible symptoms and candidate causes synthesised across the grouped records. Schema constraints, deterministic exclusion and aggregation rules, and source links support auditable processing. Model-proposed labels were accepted only with model-reported \texttt{High} confidence and no human-review flag. Accepted system-code updates covered 2,178 of 2,984~targeted records; accepted failure-mode labels covered 11,661~records (71.5\%). The enriched fields enable stratification by specific maintenance actions and failure modes, separate pitch interventions from broader system categories, and distinguish electrical and hydraulic pitch maintenance patterns. Recorded task durations totalled 6.83~hours. A source-based assessment covered 800~records selected by balanced probability sampling. Changed system codes and maintenance types required source support; retained specific labels were approved unless contradicted. Equal-weight mean approval was 86.3\% across the four fields: 94.8\% for system codes, 87.3\% for maintenance types, 85.6\% for actions, and 77.5\% for failure-mode outcomes. Failure-mode assessment covered specific titles and unresolved, unmatched or inapplicable statuses. Qualitative reviews of selected maintenance-action labels and failure-mode profiles identified unsupported actions and unclear category boundaries. The resulting dataset provides traceable evidence for engineering review and future operational linkage supporting event reconstruction, exposure-based reliability analysis, and failure modes and effects analysis.
\end{abstract}

\vspace{0.35cm}


% ==============================================================================
% NOMENCLATURE
% ==============================================================================
\section*{Nomenclature}
\label{sec:nomenclature}

\begin{tabularx}{\linewidth}{@{}>{\bfseries}l >{\RaggedRight\arraybackslash}X@{}}
API & Application Programming Interface \\
CMMS & Computerised Maintenance Management System (software for managing work orders) \\
FMEA & Failure Modes and Effects Analysis \\
LLM & Large Language Model \\
NLP & Natural Language Processing \\
RDS-PP & Reference Designation System for Power Plants (the utilised turbine-system taxonomy) \\
SCADA & Supervisory Control and Data Acquisition \\
\end{tabularx}

\newpage

% ==============================================================================
% MAIN BODY
% ==============================================================================

% ==============================================================================
% SECTION 1: INTRODUCTION
% ==============================================================================
\section{Introduction}
\label{sec:introduction}

% ------------------------------------------------------------------------------
% SUBSECTION 1.1: BACKGROUND
% ------------------------------------------------------------------------------
\subsection{Background}
\label{subsec:background}

Expanding wind generation and ageing fleets increase the importance of reliability engineering and maintenance optimisation. Operation and maintenance expenditure contributes materially to lifecycle cost, while failures can cause repair expenditure, production loss, and logistical delays~\cite{alhmoud_review_2018}. Historical operational records can provide evidence of degradation and maintenance resource use.

Reliability analysis uses historical failure observations and operating exposure, such as time in service, to estimate failure rates and downtime. Wind energy studies have applied this approach to offshore and onshore fleets. Reported results vary with population, event definition, data collection, and system taxonomy, the classification of equipment into systems and subsystems~\cite{carroll_failure_2016,lin_fault_2016,artigao_failure_2021}. Maintenance records can contribute evidence to this process, provided that recorded interventions are structured consistently and subsequently linked to operating and downtime information.

Failure modes and effects analysis (FMEA) additionally requires a defensible account of the failed item, failure mode, mechanism, effect, and existing controls~\cite{scheu_systematic_2019}. Defining modes and risk parameters relies substantially on expert judgement. Fuzzy methods, cost-weighted formulations, and modified risk models address aspects of this subjectivity~\cite{dinmohammadi_fuzzy-fmea_2013,tazi_using_2017,li_improved_2022}. Pitch-system studies use multi-criteria methods to organise heterogeneous consequences and risk judgements across interacting mechanical, electrical, and control functions~\cite{wang_integrated_2022}. These methods formalise assessment, while traceable failure-mode evidence remains necessary. Both fleet reliability analysis and FMEA therefore depend on consistent system classifications and structured diagnostic evidence.

Computerised maintenance management systems (CMMSs) contain such evidence in work-order fields and technician notes. Earlier studies used computational language analysis, known as natural language processing (NLP), to standardise maintenance text and support reliability indicators~\cite{salo_work_2019,lutz_digitalization_2022,lutz_kpi_2023,walgern_impact_2024}. Rule-based methods apply explicit patterns or mappings, whereas supervised classifiers learn from labelled examples~\cite{salo_work_2019,lutz_kpi_2023}. Both require defined target categories and assessment of their application to another dataset~\cite{walgern_impact_2024}.

More recent work has investigated large language models (LLMs) for repair recommendations and maintenance labelling through dynamically supplied schemas specifying output fields and formats~\cite{walker_using_2024,walshe_automatic_2025}.

Stewart et al.~\cite{stewart_large_2023} fine-tuned \texttt{GPT-3.5}, further training it on expert-labelled observations to classify failure modes. The $F_1$ score combines precision, the proportion of predicted labels that are correct, and recall, the proportion of reference labels recovered. Micro averaging pools label decisions, while macro averaging gives each class equal weight. Their preprint reported results for 62~test observations: micro-$F_1$ increased from 0.46 to 0.81 after fine-tuning, and the fine-tuned model's macro-$F_1$ was 0.62. Ding et al.~\cite{ding_large_2026} instead used LLM-generated minority-class examples to train a hierarchical classifier for offshore maintenance records. Its mean macro-$F_1$ was 0.821, compared with 0.704 for the strongest resampling baseline.

Youn et al.~\cite{youn_linear_2025} applied \texttt{GPT-4o-mini} to clinical linear-accelerator service reports. Two specialists independently labelled 48~randomly sampled reports for comparison. The single-institution study reported model accuracies of 0.90 for actions and 0.95 for affected subsystems, using an incompletely specified multi-label accuracy measure.

Men et al.~\cite{men_interpretable_2025} constrained diagnostic generation using a knowledge graph, a representation of entities and their relationships derived from wind-turbine fault records. Their cause-generation evaluation assessed similarity to the causes recorded in complete, non-duplicate fault records.

The preceding comparative study benchmarked models on maintenance-record tasks with predefined labels, measuring reference-label agreement, model-reported confidence, speed, and cost~\cite{malyi_comparative_2026}. A subsequent exploratory study examined taxonomy discovery and temporal, comparative, and data-quality interpretations from record batches~\cite{malyi_exploratory_2026}.

The present production workflow retains request coordination and output controls from the comparative study and taxonomy discovery from the exploratory study. It separates taxonomy proposals, record-level labelling, and aggregation. Records grouped by system and hardware configuration (topology) supply candidate action and failure-mode taxonomies. Individual requests apply these under structured-output and review rules. Software applies predefined rules (deterministic processing) to enforce exclusions and calculate distributions, while source links provide record-level provenance (Section~\ref{sec:methodology}).

% ------------------------------------------------------------------------------
% SUBSECTION 1.2: PROBLEM STATEMENT
% ------------------------------------------------------------------------------
\subsection{Problem statement}

Legacy maintenance records contain information that is not readily available in the structured form required by reliability analysis and FMEA. Work orders document interventions. Establishing failure events, physical causes, downtime, and operating exposure draws on corroborating sources such as supervisory control and data acquisition (SCADA) telemetry and alarm and status logs~\cite{malyi_wind_2025}. Before such linkage is possible, the maintenance record must identify the affected system and preserve the available evidence about the intervention, observed condition, and possible failure mode.

This gap limits fleet-level comparability and event reconstruction. Missing codes, broad classifications, generic action labels, and observations dispersed across work-log entries require substantial manual interpretation across a multi-year archive. A scalable process must retain the original evidence, expose uncertainty, and permit human or multi-source verification.

CMMS data quality issues include unstructured text, semantic ambiguity, incomplete categorical fields, and inconsistent hierarchical codes~\cite{leahy_issues_2019}. In the dataset considered here, these issues appear in five forms:
\begin{enumerate}
    \item \textbf{Missing or inconsistent system codes:} Records may omit a hierarchical code or assign a code that is not supported by the accompanying text.
    \item \textbf{Insufficient resolution in legacy system coding:} The Reference Designation System for Power Plants (RDS-PP) provides the baseline functional hierarchy~\cite{wilkinson_european_2010,VGB_RDSPP_Part32_2021}. In this portfolio, legacy codes grouped pitch interventions under rotor-blade or central-hydraulic headings, obscuring pitch as a distinct analytical group (Section~\ref{sec:methodology}). The required harmonisation addresses local recording and mapping practice. Related work shows that organisational taxonomy choices can affect subsystem reliability measures~\cite{walgern_impact_2024}.
    \item \textbf{Unstructured free-text descriptions:} Technician notes contain variations in terminology, formatting artefacts, and iterative updates that are not represented in the categorical fields.
    \item \textbf{Inconsistent maintenance-type labels:} A recorded maintenance type may differ from the activity described in the text.
    \item \textbf{Generic or missing action labels:} An action field may record only a broad verb, such as repair, without identifying the affected component or specific intervention.
\end{enumerate}

Table~\ref{tab:legacy_sample} illustrates these issues in two legacy records.

\begin{table}[!ht]
    \centering
    \scriptsize
    \setlength{\tabcolsep}{4pt}
    \renewcommand{\arraystretch}{1.1}

    \caption[Examples of legacy maintenance-record quality issues]{Source categorical fields and free-text excerpts from two legacy records in the 16,316-record dataset, illustrating missing, generic, or inconsistent entries.}
    \label{tab:legacy_sample}

    \newcolumntype{Y}{>{\RaggedRight\arraybackslash}X}

    \begin{tabularx}{\linewidth}{@{} >{\RaggedRight\arraybackslash}p{1.8cm} >{\RaggedRight\arraybackslash}p{2.2cm} >{\RaggedRight\arraybackslash}p{1.8cm} Y @{}}
        \toprule
        \textbf{System\newline Name} & \textbf{Maintenance\newline Type} & \textbf{Action\newline Taken} & \textbf{Descriptors} \\
        \midrule

        Hydraulic Systems & Corrective & [~Empty~] & Replacement of the degraded pads of the main brake disc of the drive train. \\
        \addlinespace

        [~Empty~] & Corrective & Repair & [~OEM~name~] conducted inspection to verify piping of the lubrication system interconnecting the Cooler and the Gearbox. Execution of visual inspection on corrosions. [~OEM~name~] carried out the corrosion removal work. \\

        \bottomrule
    \end{tabularx}

    \vspace{2pt}
    \begin{minipage}{\linewidth}
        \scriptsize
        \textit{Note:} OEM denotes original equipment manufacturer.
    \end{minipage}
\end{table}

The workflow organises maintenance text into traceable semantic fields for engineering review, distinguishing recorded observations, model interpretations, workflow acceptance, and later physical corroboration.

% ------------------------------------------------------------------------------
% SUBSECTION 1.3: RESEARCH OBJECTIVES AND SCOPE
% ------------------------------------------------------------------------------
\subsection{Research objectives and scope}
\label{subsec:research_objectives}

\noindent This study addresses the following research question:

\begin{quote}
\textit{How can a schema-constrained, topology-aware LLM workflow propose and assess semantic labels for fleet-scale wind turbine maintenance records, and what evidential and operational limitations govern the resulting data product?}
\end{quote}

\noindent The workflow has five objectives:
\begin{enumerate}
    \item Harmonise the legacy system taxonomy and identify records requiring system-code assignment or review.
    \item Derive candidate maintenance-action and failure-mode evidence taxonomies from system- and topology-specific record batches.
    \item Propose system codes, maintenance types, action labels, and failure-mode evidence fields at record level using structured outputs.
    \item Characterise automated workflow acceptance and assess final field outcomes through a source-based assessment of a balanced 800-record sample and targeted qualitative reviews.
    \item Report the application programming interface expenditure, runtime, throughput, and token volume of the implemented workload.
\end{enumerate}

The empirical contribution comprises fleet-scale implementation, field-level approval estimates and sampling precision, and workload measurement. The labels support case selection, provenance-supported expert audit, and subsequent operational linkage for reliability analysis and FMEA.

% ==============================================================================
% SECTION 2: METHODOLOGY
% ==============================================================================
\section{Methodology}
\label{sec:methodology}

% ------------------------------------------------------------------------------
% SUBSECTION 2.1: DATA DESCRIPTION
% ------------------------------------------------------------------------------
\subsection{Data description}
\label{subsec:data_description}

The study dataset comprises 16,316~unique maintenance records from 280~wind turbines across 31~onshore wind farms. The fleet has a total installed capacity of 635.2~MW and contains 12~turbine models with rated capacities from 0.66 to 3.2~MW. In 2026, turbine ages are concentrated between 17 and 22~years. Of the 16,316~records, 15,009~document activities assigned to wind turbine generators and 1,307~concern balance-of-plant infrastructure. The archive spans 9.16~years, from 4~January 2017 to 4~March 2026.

The source query retained work orders carrying the legacy maintenance-type labels \emph{Corrective} or \emph{Predictive}. Subsequent inspection identified records whose free-text descriptions instead supported preventive, inspection, reset, or retrofit categories. These differences between the retrieval labels and the narratives motivate maintenance-type relabelling.

% ------------------------------------------------------------------------------
% SUBSECTION 2.2: TERMINOLOGY
% ------------------------------------------------------------------------------
\subsection{Terminology}
\label{sec:terminology}

The source CMMS is IBM Maximo~\cite{ibm_maximo_2024}. A \textit{work order} authorises and records an intervention; a \textit{work log} is an associated timestamped textual update. A \textit{maintenance record}, or \textit{record}, combines retained work-order attributes and chronologically ordered work-log text in one analytical row per unique work-order identifier.

The original dataset uses the RDS-PP taxonomy~\cite{wilkinson_european_2010,VGB_RDSPP_Part32_2021}. A \textit{system code} is a hierarchical designation for the functional system or subsystem to which an intervention is assigned. A \textit{component} is a more specific physical item within that hierarchy and may fall below the available RDS-PP code granularity. Component information is consequently retained from source fields where available or proposed from the maintenance text by the workflow.

A \textit{failure-mode evidence profile} comprises a failure-mode title, common mechanisms, possible symptoms, and candidate causes. The title describes the condition assigned to the item. Common mechanisms are representative explanations of how it may develop, synthesised from maintenance evidence and relevant domain knowledge. Possible symptoms describe indications associated with the profile, and candidate causes propose initiating or contributing explanations. These fields provide profile-level engineering context for records receiving the label.

A \textit{candidate functional-failure record} is retained for descriptive analysis and engineering review by a deterministic rule combining maintenance type, action, and failure-mode outcomes (Section~\ref{subsubsec:candidate_record_selection}). An informative action can support inclusion for selected maintenance types even when failure mode is unresolved.

In future applications, an explicit work-order eligibility rule would define a \textit{maintenance event}. An \textit{operational episode} would denote a bounded period of non-generation reconstructed from operational sources. These units distinguish documented maintenance activity from reconstructed operating states.

% ------------------------------------------------------------------------------
% SUBSECTION 2.3: DATA PROCESSING AND AUTOMATED LABELLING PIPELINE
% ------------------------------------------------------------------------------
\subsection{Data processing and automated labelling pipeline}
\label{subsec:labelling_pipeline}

Figure~\ref{fig:master_workflow} connects database preparation and taxonomy harmonisation with semantic labelling, output controls, action refinement, and programmatic aggregation~\cite{malyi_2026_zenodo}. Section~\ref{subsubsec:taxonomy_sequence} details the task sequence; Section~\ref{subsubsec:acceptance_rules} defines acceptance and deterministic bypass rules.

\begin{figure}[p]
    \centering
    \scalebox{0.90}{\input{Figure_1}}
    \caption[Overview of the maintenance-record labelling workflow]{Overview of the maintenance-record labelling workflow. Solid arrows show processing and outputs; dashed arrows show contextual inputs. CMMS denotes computerised maintenance management system; RDS-PP denotes Reference Designation System for Power Plants.}
    \label{fig:master_workflow}
\end{figure}

% ..............................................................................
% SUBSUBSECTION 2.3.1: DATABASE INTEGRATION AND PRE-PROCESSING
% ..............................................................................
\subsubsection{Database integration and pre-processing}
\label{subsubsec:database_preprocessing}

Database tables were linked using the original work-order number as a stable key. The \texttt{WORKORDER} table supplied one row and its primary description for each work order. The \texttt{WORKLOG} table supplied zero or more timestamped technician notes, while \texttt{FAILURECODE} supplied the description associated with the recorded failure code.

Linking each row in \texttt{WORKORDER} to its associated \texttt{WORKLOG} entries initially repeated work-order identifiers. The linked rows were therefore regrouped by unique work-order number to recover one analytical maintenance record for each of the 16,316~work orders. Identical work-log strings attached to the same work order were suppressed, and distinct retained notes were ordered by their timestamps. Primary descriptions, additional observations, chronological work logs, and failure-code descriptions were combined into \texttt{Merged\_Descriptors} (\texttt{Descriptions} in the study export) using source-labelled separators. The work-order identifier, retained as \texttt{WO\_ID}, links \texttt{Merged\_Descriptors} to the corresponding source CMMS records for audit.

Predefined text-matching rules (regular expressions) were then applied to the unique merged text strings. They removed database line breaks, markup tags, escape sequences, and repeated formatting characters. They also masked commercially sensitive asset, contractor, and site identifiers and normalised a defined set of recurrent abbreviations.

Later work-log entries can add diagnostic observations, replaced components, and restoration actions to a generic primary description. Chronological consolidation preserves this documentary sequence, including administrative updates.

% ..............................................................................
% SUBSUBSECTION 2.3.2: TAXONOMY ALIGNMENT AND TASK SEQUENCE
% ..............................................................................
\subsubsection{Taxonomy alignment and task sequence}
\label{subsubsec:taxonomy_sequence}

The RDS-PP taxonomy, including its wind-energy adaptation, provided the baseline system hierarchy~\cite{wilkinson_european_2010,VGB_RDSPP_Part32_2021}. A predefined dictionary mapping harmonised 123~active legacy codes observed in the portfolio into 99~target codes before model processing. Legacy codes comprised 75~wind-turbine and 48~balance-of-plant codes; target codes comprised 63 and 36, respectively. Separating \emph{Pitch System} expanded the wind-turbine grouping from 18~legacy groups to 19~target groups. The target taxonomy comprises 24~functional groups: 19~wind-turbine and 5~balance-of-plant groups. Mapping also consolidated duplicated cooling or switchgear categories and restored missing dictionary descriptions where a source reference was available, retaining original codes and mapping provenance.

The subsequent semantic workflow comprised five main task types:
\begin{enumerate}
    \item \textbf{System-code classification:} Records with missing codes, candidate legacy pitch-category discrepancies, or audit flags were processed against the applicable target taxonomy to propose a system code and component.
    \item \textbf{System-code audit:} The model compared the current system assignment with the merged record text and returned a binary relabelling flag with an evidence rationale, confidence tier, and review flag.
    \item \textbf{Batch extraction of candidate taxonomies:} Records were grouped by assigned system and relevant topology to propose candidate action taxonomies and failure-mode evidence profiles. Each populated system-code group was processed as a separate extraction batch, with further subdivision where applicable topology configurations differed. The prompt requested retention of rare, technically specific cases, including single-occurrence anomalies. It also requested separate profiles where distinct mechanisms or causes were inferred~\cite{malyi_2026_zenodo}.
    \item \textbf{Maintenance-type and action labelling:} Each record was processed in a separate request against the six-class maintenance-type taxonomy (\emph{Corrective}, \emph{Reset}, \emph{Preventive}, \emph{Predictive}, \emph{Inspection}, and \emph{Retrofit}) and its system-specific candidate action labels.
    \item \textbf{Failure-mode labelling:} Records requiring inference were processed in separate requests against the system- and topology-specific failure-mode evidence dictionary.
\end{enumerate}

The six maintenance types were specified before record-level labelling as mutually exclusive analytical labels for this study. Their definitions drew on EN~13306 terminology with study-specific customisation~\cite{en_maintenance_terminology_2017}. The standard treats predictive maintenance as a form of condition-based preventive maintenance and inspection as an activity. The prompt uses \emph{Preventive} for routine calendar- or usage-based servicing, \emph{Reset} for restoration through control or software actions, and \emph{Retrofit} for campaigns, upgrades, or design modifications. \emph{Inspection} denotes checks without immediate repair, whereas \emph{Predictive} denotes targeted work prompted by an anticipated failure. Reported maintenance-type outcomes additionally include \texttt{Other} and \texttt{Not enough information}.

Targeted missing-code and pitch records bypass the broader audit. Missing-code recovery begins without a usable assignment; pitch relabelling begins with deterministic text and taxonomy evidence of a known discrepancy. For remaining records, the audit tests the existing assignment before routing flagged records to classification. The two classification boxes in Figure~\ref{fig:master_workflow} represent these distinct uses of the same task type.

Each task used labels accepted or retained at earlier stages. System codes determined how records were grouped for candidate taxonomy extraction and which system-specific context and dictionaries were supplied to record-level requests. Maintenance-type and action labelling interpreted the merged narrative within that context. The resulting maintenance type governed deterministic failure-mode exclusions; eligible requests also received maintenance type and action alongside the applicable failure-mode dictionary~\cite{malyi_2026_zenodo}.

Following initial failure-mode labelling, records with a specific failure-mode label and \texttt{Other} in the maintenance-action field entered unexpected-action refinement. Semantic clustering grouped these action descriptions by meaning to propose refined action categories. A subsequent record-level action-relabelling pass applied the refined categories to the selected records~\cite{malyi_2026_zenodo}.

Archive-derived taxonomies capture portfolio-specific distinctions in shared categories. Python calculated counts and proportions from final record-level labels and workflow statuses. For sparse records, common mechanisms and candidate causes could be inherited from the applicable evidence dictionary.

\afterpage{\clearpage}

% ------------------------------------------------------------------------------
% SUBSECTION 2.4: DESIGN CONSIDERATIONS AND DECISIONS
% ------------------------------------------------------------------------------
\subsection{Design considerations and decisions}

% ..............................................................................
% SUBSUBSECTION 2.4.1: MODEL CONFIGURATION AND PROCESSING ARCHITECTURE
% ..............................................................................
\subsubsection{Model configuration and processing architecture}
\label{subsubsec:model_configuration}

The workflow used OpenAI's proprietary \texttt{gpt-5.4-thinking} model~\cite{openai_gpt54_2026} throughout the reported pipeline. It was accessed through the OpenAI application programming interface (API) in Python~\cite{openai_python_2025} over multiple dates in April~2026. Prompt assembly, request coordination (orchestration), and schema validation were separated from the model interface. This model-agnostic design permits another compatible model to be evaluated without rewriting the complete workflow.

The production model was selected after the earlier comparative and exploratory studies had been completed~\cite{malyi_comparative_2026,malyi_exploratory_2026}. Selection reflected institutional availability, support for structured outputs and configurable reasoning effort, and the provider's reported performance across broad benchmarks~\cite{openai_gpt54_2026}.

Medium reasoning effort was used for system-code auditing, record-level classification, and taxonomy labelling; High effort was used for multi-record batch extraction. Temperature was unavailable for this configuration and was not specified. Each record-level request used a fresh conversation with the applicable taxonomy and evidence dictionary supplied explicitly. Asynchronous queues capped simultaneous requests at 10~for record-level tasks and 5~for batch extraction, whose contexts and outputs were larger.

% ..............................................................................
% SUBSUBSECTION 2.4.2: TOPOLOGY-AWARE PROMPTING AND SCHEMA VALIDATION
% ..............................................................................
\subsubsection{Topology-aware prompting and schema validation}
\label{subsubsec:topology_prompting}

Prompt development included iterative tests of wording and different reasoning-effort settings on smaller samples of maintenance records. Outputs, including the generated candidate taxonomies, were reviewed qualitatively for adherence to the instructions and the required response structure.

The final automated run proceeded from batch taxonomy extraction to record-level labelling without an intermediate expert review of the candidate taxonomies. The balanced field-level assessment and qualitative reviews were performed after labelling was complete (Sections~\ref{subsec:validation_design} and~\ref{subsec:semantic_label_review}).

Each prompt included the applicable system taxonomy and relevant hardware configuration details:
\begin{enumerate}
    \item \textbf{Generator topology:} Squirrel-cage induction, wound-rotor induction, doubly-fed induction, or direct-drive electrically excited synchronous generator, with the corresponding generator and converter configuration.
    \item \textbf{Drivetrain topology:} High-speed geared or direct-drive architecture.
    \item \textbf{Pitch-system architecture:} Electromechanical pitch drives with motors and backup batteries, or hydraulic systems with cylinders and accumulators.
    \item \textbf{Yaw-mechanism architecture:} Electromechanical yaw drives, hydraulic friction callipers, or passive brake assemblies.
\end{enumerate}

Where topology metadata were unavailable or inapplicable to the assigned system, prompts retained the system-specific context without a topology qualifier.

Topology flags narrow candidate sets to the recorded hardware. For example, electrical pitch systems use motors, inverters, chargers, and batteries, whereas hydraulic systems use pumps, valves, cylinders, and accumulators. This context constrains interpretations of generic terms such as drive or controller.

Responses were checked against task-specific JSON schemas for structure and required fields. Unparseable or incomplete responses triggered repeat requests. No requests remained unanswered or structurally invalid after retries.

% ..............................................................................
% SUBSUBSECTION 2.4.3: EVIDENCE RATIONALES, CONFIDENCE TIERS, AND DETERMINISTIC RULES
% ..............................................................................
\subsubsection{Evidence rationales, confidence tiers, and deterministic rules}
\label{subsubsec:acceptance_rules}

Record-level labelling and audit prompts requested a one-sentence evidence rationale linking each proposed output to the source text for inspection.

Each response included a model-generated confidence tier (\texttt{High}, \texttt{Medium}, or \texttt{Low}) and human-review flag for review routing. Confidence is an ordinal self-assessment, not a calibrated probability of correctness. Across record-level labelling tasks, an output is reported as accepted only with \texttt{High} confidence and no human-review flag. Within the combined maintenance-type/action request, each field was assessed separately against this rule. Automated acceptance determines inclusion in label summaries and is distinct from source-based assessment approval (Sections~\ref{subsec:validation_design} and~\ref{subsec:validation_results}). Deterministic exclusions and other workflow statuses are reported separately.

In system-code processing, an accepted proposal updated the existing system code or populated a missing code. A proposal that failed either acceptance criterion was discarded, leaving the existing code or missing-code status unchanged.

For maintenance type, action, and failure mode, a proposal that failed either acceptance criterion resulted in \texttt{Not enough information} for that field. The model could also return this status directly. Unresolved totals include both routes and, for failure modes, deterministic propagation from unresolved actions.

Before failure-mode inference, deterministic processing first checked whether the action was \texttt{Not enough information}. If so, the failure-mode field received the same status. Otherwise, records labelled \emph{Preventive} or \emph{Retrofit} received \texttt{Not Applicable}. This ordering gave the unresolved-action rule precedence over the maintenance-type exclusion and reduced unnecessary model calls. The failure-mode prompt could also return \texttt{Not Applicable} for descriptions interpreted as non-failure activities, including inspections.

The action and failure-mode taxonomies included \texttt{Other}, used when the model found no matching candidate label despite informative text.

% ..............................................................................
% SUBSUBSECTION 2.4.4: CANDIDATE-RECORD SELECTION AND SOURCE REVIEW
% ..............................................................................
\subsubsection{Candidate-record selection and source review}
\label{subsubsec:candidate_record_selection}

Figures~\ref{fig:Figure_11}, \ref{fig:Figure_12}, and~\ref{fig:Figure_13} use a candidate-record flag calculated from \texttt{Maintenance\_Type}, \texttt{Action\_Taken}, and \texttt{Failure\_Mode}. A flag of \texttt{True} retains a record under the following rules:
\begin{enumerate}
    \item For \emph{Corrective}, \emph{Predictive}, or \emph{Reset}, retain the record unless both action and failure mode have outcomes drawn from \{\texttt{Not enough information}, \texttt{Not Applicable}\}. Either field can therefore support inclusion; \texttt{Other} is retained as an informative unmatched outcome.
    \item For \emph{Inspection} or \texttt{Other} maintenance types, retain the record only when failure mode is neither \texttt{Not enough information}, \texttt{Not Applicable}, nor \texttt{Other}.
    \item Exclude all remaining maintenance types, including \emph{Preventive}, \emph{Retrofit}, and \texttt{Not enough information}.
\end{enumerate}

This selection rule is separate from the automated acceptance of a proposed label. It identifies a descriptive population whose actions and possible failure conditions warrant review; it does not confirm independent physical failures.

To examine unresolved outcomes, records with \texttt{Not enough information} were tabulated by system group and maintenance type. Descriptor length was measured by counting whitespace-separated words in \texttt{Descriptions} for all 3,441~records with unresolved failure modes. Qualitative source review covered all 2,065~records unresolved in maintenance type, action, and failure mode, and all 1,214~records with \texttt{Not Applicable} failure-mode outcomes. These readings examined the information available for labelling and whether descriptions excluded from failure-mode labelling contained failure-related evidence. They were descriptive source reviews, separate from the sampled binary assessment in Section~\ref{subsec:validation_design}.

% ------------------------------------------------------------------------------
% SUBSECTION 2.5: DATA INPUT AND OUTPUT STRUCTURE
% ------------------------------------------------------------------------------
\subsection{Data input and output structure}

Table~\ref{tab:input_output_parameters} summarises inputs, outputs, and controls for the five main task types. Subsequent action refinement is detailed in Section~\ref{subsubsec:taxonomy_sequence}.

\begin{table}[!ht]
    \centering
    \footnotesize
    \setlength{\tabcolsep}{4pt}
    \renewcommand{\arraystretch}{1.2}
    \caption[Inputs, outputs, and controls for the five main pipeline task types]{Inputs, model-generated outputs, and output controls for the five main pipeline task types. Subsequent action refinement is shown separately in Figure~\ref{fig:master_workflow}. The controls support schema conformance and review routing.}
    \label{tab:input_output_parameters}

    \begin{tabularx}{\linewidth}{@{} >{\RaggedRight\arraybackslash}p{2.8cm} >{\RaggedRight\arraybackslash}X >{\RaggedRight\arraybackslash}X >{\RaggedRight\arraybackslash}p{3.2cm} @{}}
        \toprule
        \textbf{Task Type} & \textbf{Input Parameters} & \textbf{Generated Outputs} & \textbf{Output Controls} \\
        \midrule
        System-code classification & Merged record text, current code and component, topology, applicable RDS-PP taxonomy & Proposed system code and component & Evidence rationale, confidence tier, human-review flag \\
        \addlinespace
        System-code audit & Merged record text, current code and component, topology & Binary relabelling flag & Evidence rationale, confidence tier, human-review flag \\
        \addlinespace
        Candidate taxonomy extraction & Component, topology, batch of merged record text & Candidate maintenance actions and failure-mode evidence profiles & Schema-constrained candidate labels; batch estimates excluded from quantitative results \\
        \addlinespace
        Maintenance-type and action labelling & Merged record text, component, topology, maintenance-type taxonomy, action taxonomy & Proposed maintenance type and action label & Evidence rationale, confidence tier, human-review flag \\
        \addlinespace
        Failure-mode labelling & Merged record text, component, topology, maintenance type, action, failure-mode evidence dictionary & Proposed failure-mode title, common mechanisms, possible symptoms, and candidate causes & Evidence rationale, confidence tier, human-review flag \\
        \bottomrule
    \end{tabularx}
\end{table}

% ------------------------------------------------------------------------------
% SUBSECTION 2.6: VALIDATION DESIGN AND ASSESSMENT
% ------------------------------------------------------------------------------
\subsection{Validation design and assessment}
\label{subsec:validation_design}

% ..............................................................................
% SUBSUBSECTION 2.6.1: SUBSET SELECTION
% ..............................................................................
\subsubsection{Subset selection}

The validation subset comprised 800~unique records selected without replacement from the 16,316-record archive (4.90\%), covering specific labels and escape outcomes (insufficient information, no suitable category, or inapplicability). Each record contributed four assessment decisions irrespective of candidate-record eligibility. The sample size balanced review resources with a worst-case 95\% sampling margin of approximately $\pm3.4$~percentage points under simple random sampling with a finite-population correction. This planning reference differs from the balanced-sampling approximation used for the reported intervals.

Selection used equal-probability balanced sampling based on the cube method~\cite{deville_cube_2004}. The design sought to reproduce the archive proportions separately for system-code processing, maintenance-type outcomes, action outcomes, and failure-mode outcomes. Their 15~category indicators and an intercept fixing sample size formed a balancing matrix of rank~12. With $n=800$ and $N=16{,}316$, each record had inclusion probability $\pi_i=n/N\approx0.04903$ and expansion weight $N/n=20.395$. For a category containing $N_c$~archive records, the proportional sample target was $t_c=nN_c/N$; the achieved integer counts approximated these targets.

System-code processing distinguished accepted updates by correction workflow from retained codes or missing-code statuses. Changed and retained maintenance types covered named categories only, with \texttt{Other} and \texttt{Not enough information} kept separate regardless of their legacy values. Action and failure-mode escape outcomes combined model-returned statuses with those applied after failed acceptance or through deterministic processing.

In the flight stage of the cube method, inclusion probabilities moved towards selection or exclusion while preserving the balancing totals. The pairwise pivotal landing stage resolved the remaining fractional probabilities and fixed the sample at 800~records. Pairs were chosen to minimise the expected increase in summed squared marginal count errors, allowing small departures from category targets. Both stages preserved each record's inclusion probability in expectation. The design and seed were specified before selection, and archive positions were randomly permuted before a single draw. Diagnostics prompted neither redraws nor record swaps. Assessment order was randomised separately without changing sample membership.

Post-selection diagnostics compared system groups, topology within relevant systems, action-start years, date ranges, and description lengths using the original archive fields. These additional characteristics did not constrain selection.

% ..............................................................................
% SUBSUBSECTION 2.6.2: ASSESSMENT RULES
% ..............................................................................
\subsubsection{Assessment rules}

The first author assessed the final system-code, maintenance-type, action, and failure-mode outcomes separately against each record's complete merged source narrative and supplied equipment metadata. Each decision was binary: approval indicated that the applicable source-support or unchanged-label retention criterion was met; rejection indicated that it was not. Assessment approval was distinct from automated workflow acceptance (Section~\ref{subsubsec:acceptance_rules}). Concise rejection reasons were recorded, with supporting excerpts and alternative labels where available for a downstream qualitative overview.

Changed specific system codes and maintenance types required support at the stated specificity. Unchanged specific labels were approved unless the source clearly contradicted them; insufficient text alone did not warrant rejection. Approval in these retained groups therefore indicated that no clear discrepancy had been found, rather than necessarily confirming the label from the narrative. The retention criterion did not apply to escape outcomes.

System-code assessment concerned the affected component's functional system, allowing a parent-system code at the intended taxonomy depth. Maintenance-type assessment concerned intervention purpose across the complete account under the definitions in Section~\ref{subsubsec:taxonomy_sequence}; inspection followed by repair was not treated as inspection-only work. An explicit recommendation could support a maintenance type when its purpose was identifiable, even if completion was unconfirmed.

Action assessment required source support for every asserted action and target; a material unsupported addition warranted rejection of a composite label. Explicit recommendations and plans could support action labels without confirmation that the work had occurred. Omission of another documented activity alone did not warrant rejection, so approval tested label support rather than exhaustive coverage.

For specific failure-mode labels, assessment covered the title only, including diagnostic details asserted within it. Approval required a source-supported engineering interpretation of the recorded condition and intervention, using the supplied equipment context and the assessor's familiarity with maintenance patterns observed in the archive. Materially conflicting interpretations and unsupported diagnostic specificity warranted rejection. Common mechanisms, possible symptoms, and candidate causes were outside this record-level binary assessment. These profile fields summarise patterns synthesised from system-specific batches of maintenance narratives drawn from the 9.2-year archive; an individual record need not document every associated mechanism, symptom, or candidate cause.

Escape outcomes were assessed under field-specific rules regardless of their processing route. An unresolved system code, stored as \texttt{Not specified}, was approved when the context could not support a system code at the applicable taxonomy depth. For other fields, \texttt{Not enough information} was approved when the evidence did not support an informative interpretation. A supported specific label or informative \texttt{Other} outcome warranted rejection of that status. \texttt{Other} was approved only when informative evidence was present without a suitable specific category.

For failure modes, source-supported non-failure work also warranted rejection of \texttt{Not enough information} in favour of \texttt{Not Applicable}. The latter was approved only for a supported non-failure activity or condition. Failure-related evidence warranted rejection of this status even when the exact mode remained unresolved. Missing causal detail alone did not justify an unresolved title when the condition could be characterised.

Rejection notes were subsequently reviewed against their supplied source context and the field-specific rules. Recurring problems were synthesised by field and outcome subgroup without quantifying the frequency of individual rejection reasons.

% ..............................................................................
% SUBSUBSECTION 2.6.3: APPROVAL ESTIMATION AND SAMPLING UNCERTAINTY
% ..............................................................................
\subsubsection{Approval estimation and sampling uncertainty}
\label{subsubsec:approval_uncertainty}

For field $j$, let $y_{ij}=1$ denote approval and $y_{ij}=0$ rejection for sampled record $i$. Equal inclusion probabilities give the archive approval estimate
\begin{equation}
    \widehat{p}_j=\frac{1}{n}\sum_{i\in s}y_{ij},
    \qquad n=800,
    \label{eq:validation_approval}
\end{equation}
where $s$ denotes the validation subset. The equal-weight mean across the four fields was calculated from each record's fraction of approved fields,
\begin{equation}
    z_i=\frac{1}{4}\sum_{j=1}^{4}y_{ij},
    \qquad
    \bar z_s=\frac{1}{n}\sum_{i\in s}z_i.
    \label{eq:validation_mean}
\end{equation}
The mean summarises approval across fields; the proportion of records with all four fields approved is a separate measure.

The pre-specified variance approximation drew on the balanced-sampling approach of Deville and Till\'{e}~\cite{deville_variance_2005}. For each field, ordinary least-squares regression of $y_{ij}$ on the sampled balancing indicators and intercept gave residuals $e_{ij}$. With $r$ denoting the rank of this sampled design matrix, the implemented approximation was
\begin{equation}
    \widehat{\operatorname{Var}}(\widehat{p}_j)
    =\frac{1-n/N}{n(n-r)}\sum_{i\in s}e_{ij}^{2}.
    \label{eq:validation_variance}
\end{equation}
Here, $1-n/N$ is the finite-population correction and $n-r$ the residual degrees of freedom. Approximate 95\% confidence intervals used $\widehat{p}_j\pm1.96\sqrt{\widehat{\operatorname{Var}}(\widehat{p}_j)}$, with endpoints restricted to $[0,1]$. The same residual procedure was applied to $z_i$ for the cross-field mean, retaining 800~records as the sampling units and preserving dependence between decisions within each record. The normal approximation requires caution near unanimous decisions or negligible residual variance. The intervals describe sampling precision under the assessment rubric.

% ==============================================================================
% SECTION 3: RESULTS
% ==============================================================================
\section{Results}
\label{sec:results}

% ------------------------------------------------------------------------------
% SUBSECTION 3.1: LABELLING RESULTS
% ------------------------------------------------------------------------------
\subsection{Labelling results}
\label{subsec:labelling_results}

Table~\ref{tab:post_processing_sample} presents the proposed semantic fields for the two legacy records introduced in Table~\ref{tab:legacy_sample}.

\begin{table}[!ht]
    \centering
    \scriptsize
    \setlength{\tabcolsep}{4pt}
    \renewcommand{\arraystretch}{1.1}

    \caption[Examples of proposed semantic enrichment]{Proposed semantic enrichment of the two legacy maintenance records reproduced in Table~\ref{tab:legacy_sample}. The model-generated fields comprise system code, maintenance type, action, and failure-mode profile descriptions.}
    \label{tab:post_processing_sample}

    \newcolumntype{Y}{>{\RaggedRight\arraybackslash}X}
    \begin{tabularx}{\linewidth}{@{} >{\RaggedRight\arraybackslash}p{1.7cm} >{\RaggedRight\arraybackslash}p{1.9cm} Y Y Y Y Y @{}}
    \toprule
    \textbf{System\newline Name} & \textbf{Maintenance\newline Type} & \textbf{Action\newline Taken} & \textbf{Failure\newline Mode} & \textbf{Common Mechanisms} & \textbf{Possible Symptoms} & \textbf{Candidate Causes} \\
    \midrule

    Drive Train Brake & Corrective & Replace brake pads & Brake pads worn or thermally degraded & Friction material wear and heat ageing reduced pad thickness and braking condition & Pads reached wear limit or temperature-related degradation and were replaced & Normal service wear accelerated by repeated braking duty and temperature exposure \\
    \addlinespace

    Gearbox Lubrication System & Inspection & Remove corrosion from cooler-to-gearbox tubing & Corrosion on cooler-to-gearbox tubing & Surface corrosion attacks external tube material and protective coating & Visual corrosion found on interconnecting tubes and corrosion removal work & Environmental exposure and coating breakdown \\

    \bottomrule
    \end{tabularx}
\end{table}

The examples recover a specific brake-system code and missing action from the first record, and identify the lubrication system and corrosion-removal action in the second. Its proposed \emph{Inspection} label is contestable because treatment followed the inspection, whereas the prompt defines inspection as work without immediate repair. More specific fields therefore remain subject to source-based review; associated mechanisms and causes provide profile-level context.

% ..............................................................................
% SUBSUBSECTION 3.1.1: SYSTEM-CODE ASSIGNMENTS AND ACCEPTANCE YIELDS
% ..............................................................................
\subsubsection{System-code assignments and acceptance yields}
\label{subsubsec:system_code_results}

\begin{figure}[!ht]
    \centering
    \includegraphics[width=\linewidth]{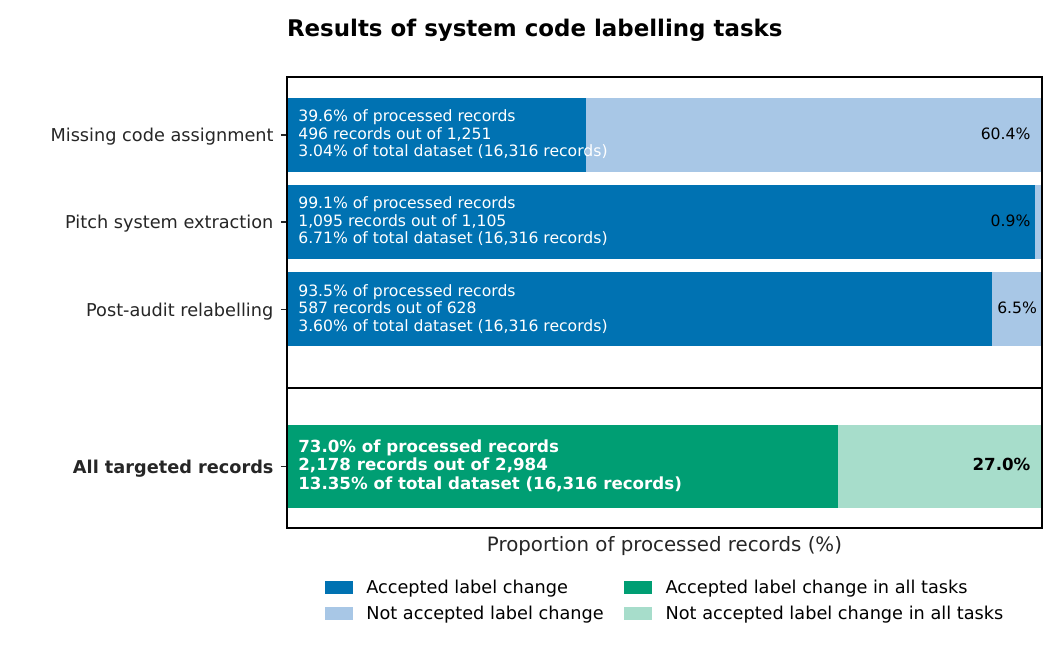}
    \caption[System-code update acceptance yields]{System-code update acceptance yields for three targeted tasks. Dark blue task segments and the dark green aggregate segment represent proposals with \texttt{High} model-reported confidence and no human-review flag. Lighter segments represent rejected updates, leaving existing codes or missing-code statuses unchanged. Task bars use their targeted records as denominators; the aggregate uses all 2,984~targeted records. Annotations also give accepted updates as shares of the full 16,316-record dataset.}
    \label{fig:Figure_2}
\end{figure}

The system-code workflow targeted 2,984~records with missing codes, candidate pitch-category discrepancies, or audit discrepancies. Figure~\ref{fig:Figure_2} reports acceptance yields under the joint rule in Section~\ref{subsubsec:acceptance_rules}. These are descriptive proportions across every record in each targeted subset during the reported execution, separate from source-based approval (Section~\ref{subsec:validation_results}).

The broader system-code audit covered 13,960~records, excluding the 1,251~records targeted for missing-code recovery and the 1,105~records targeted for pitch-category relabelling. It routed 628~records to post-audit classification; the remaining 13,332~records received no relabelling flag.

Across the three tasks, 2,178~updates were accepted (73.0\% of targeted records; 13.35\% of the corpus). Acceptance was lowest for missing-code recovery (496/1,251, 39.6\%), compared with targeted pitch relabelling (1,095/1,105, 99.1\%) and post-audit classification (587/628, 93.5\%). This variation is consistent with their input contexts: missing-code records lacked an initial system context and often had sparse text, whereas the other subsets had been screened for pitch evidence or an audit discrepancy.

\begin{figure}[!ht]
    \centering
    \includegraphics[width=\linewidth]{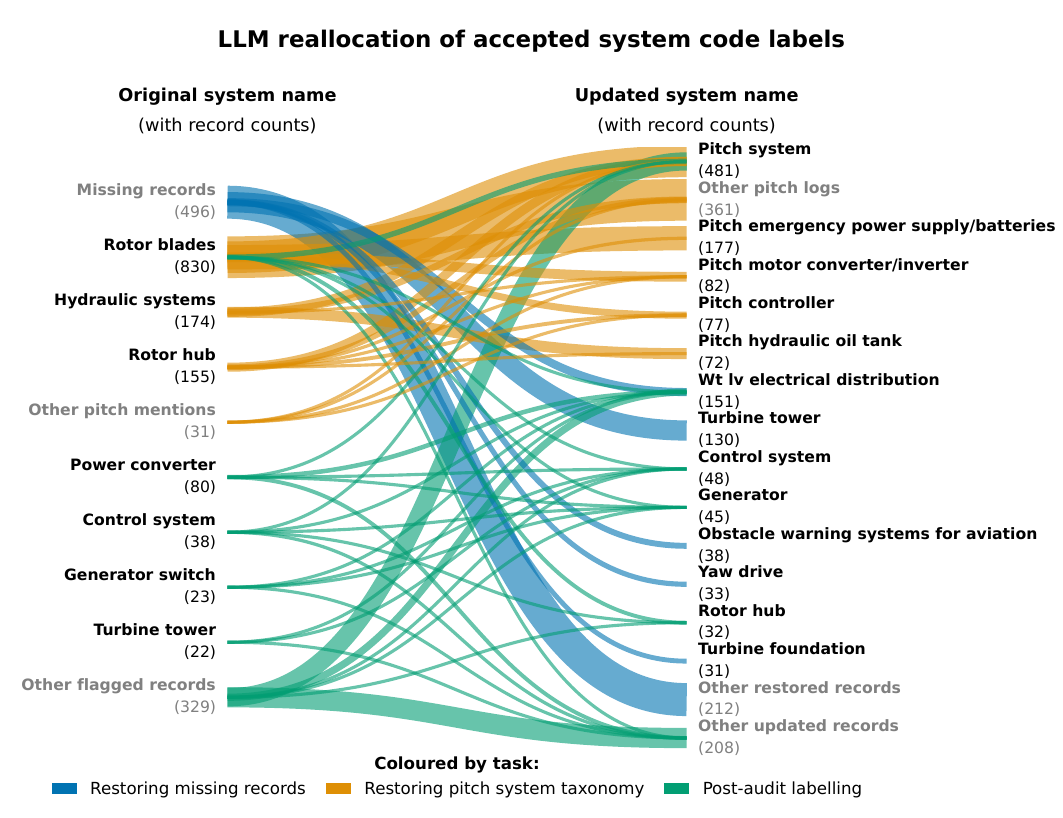}
    \caption[Accepted system-code assignment flows]{Sankey diagram from legacy to accepted model-proposed systems for all 2,178~updated records across the three targeted tasks. Widths encode record counts; colours distinguish missing-code recovery, pitch separation, and post-audit relabelling. Less frequent categories are grouped for display. WT denotes wind turbine; LV denotes low voltage.}
    \label{fig:Figure_3}
\end{figure}

Figure~\ref{fig:Figure_3} shows that the largest reallocation separates pitch interventions from rotor, hub, or hydraulic categories, allowing emergency-power, controller, and hydraulic components to be represented where supported. Missing-code recovery adds previously uncoded records to system summaries across low-voltage distribution, tower, foundation, and other systems. The more dispersed post-audit flows extend the review beyond targeted pitch work.

% ..............................................................................
% SUBSUBSECTION 3.1.2: MAINTENANCE-TYPE ASSIGNMENTS
% ..............................................................................
\subsubsection{Maintenance-type assignments}

\begin{figure}[!ht]
    \centering
    \includegraphics[width=\linewidth]{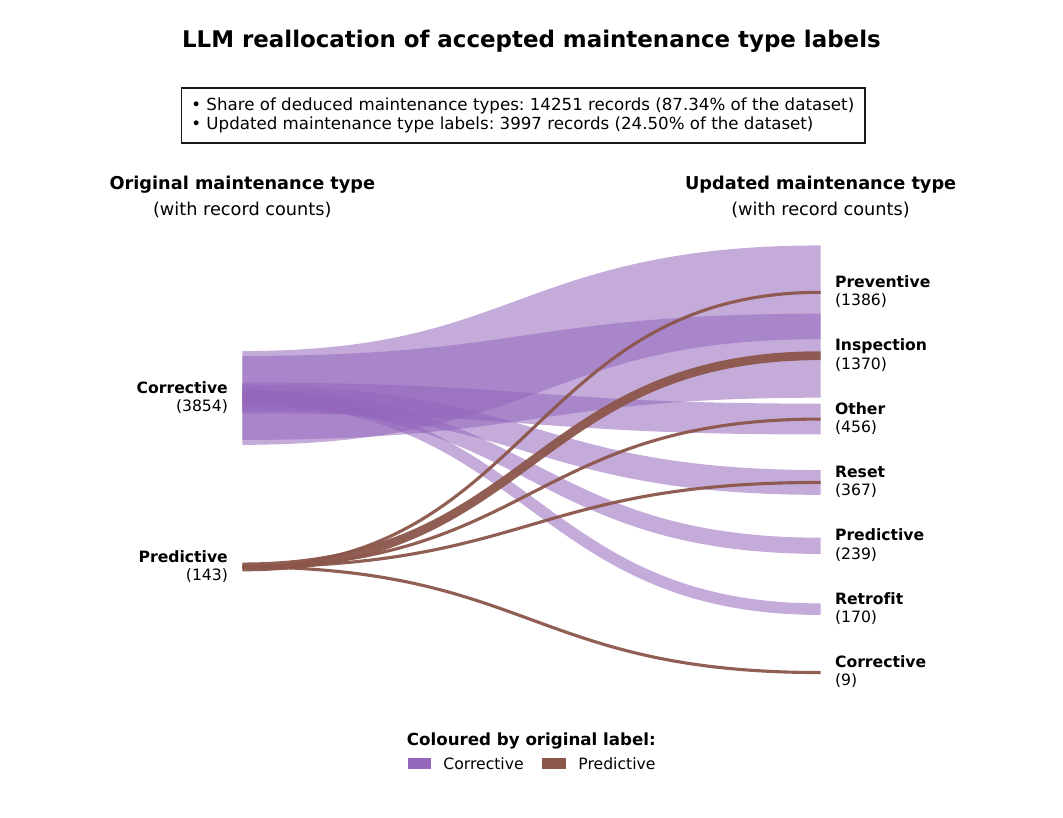}
    \caption[Accepted maintenance-type assignment changes]{Sankey diagram of original to accepted model-proposed maintenance types for 3,997~changed records (24.50\% of the 16,316-record dataset). Widths encode record counts; colours identify original labels.}
    \label{fig:Figure_4}
\end{figure}

The workflow produced accepted maintenance-type labels for 14,251~records (87.34\% of the dataset). Among them, 3,997~records, or 24.50\% of all 16,316~records, received an accepted assignment that differed from the legacy label. These comprise 3,541~records with changed named labels and 456~with \texttt{Other} outcomes, kept separate for sampling and assessment (Section~\ref{subsec:validation_results}). Figure~\ref{fig:Figure_4} shows only these differing assignments.

The principal flows move legacy corrective records to preventive maintenance, inspection, reset, and predictive maintenance, and predictive records to inspection. Distinguishing scheduled work, inspection, and restoration through resets within the two-label source query changes which records enter the candidate functional-failure set. The reallocations reflect interpretation of the merged narrative under the adopted class definitions.

% ..............................................................................
% SUBSUBSECTION 3.1.3: MAINTENANCE-ACTION ASSIGNMENTS
% ..............................................................................
\subsubsection{Maintenance-action assignments}
\label{subsubsec:maintenance_action_results}

\begin{figure}[!ht]
    \centering
    \includegraphics[width=\linewidth]{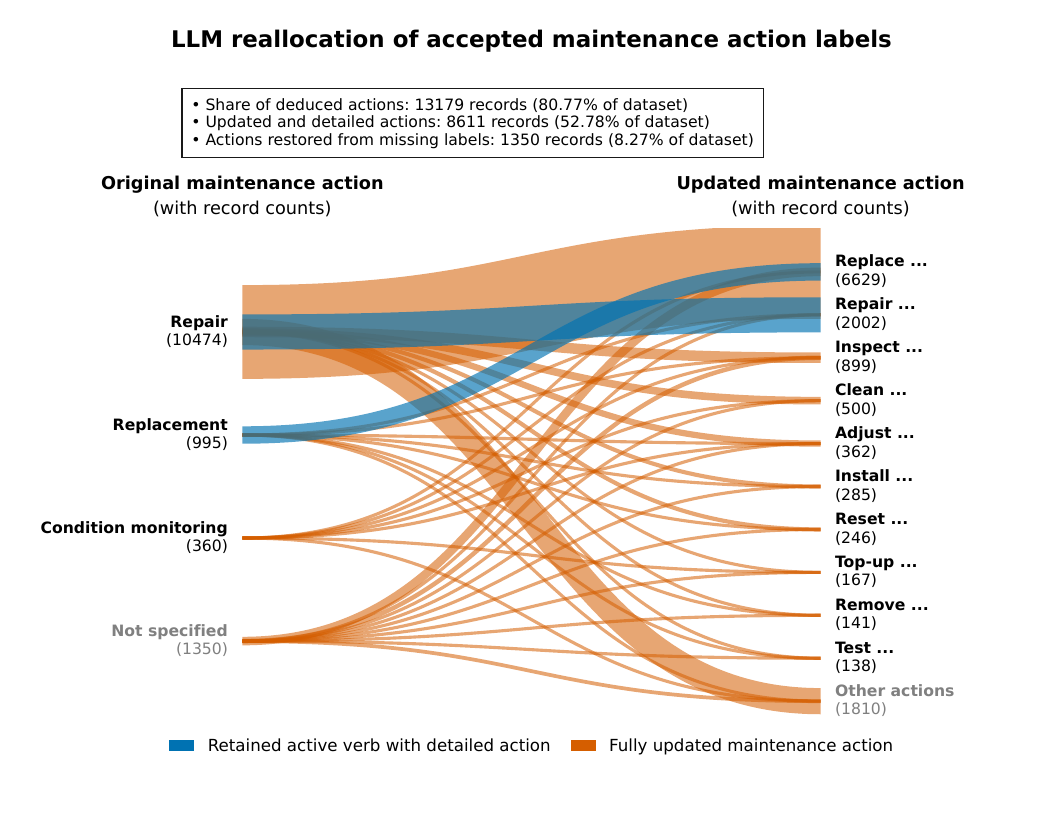}
    \caption[Accepted maintenance-action assignment changes]{Sankey diagram of original actions to model-proposed verbs for 13,179~records with accepted action labels, each contributing once. Widths encode record counts. Blue flows retain the original verb with added component or intervention detail; orange flows change the verb or populate an empty field (\emph{Not specified}). Right-hand labels group verbs; ellipses omit full-label detail. The figure includes 8,611~updated or detailed actions and 1,350~populated missing labels.}
    \label{fig:Figure_5}
\end{figure}

The workflow produced accepted maintenance-action labels for 13,179~records (80.77\% of the study dataset). Of these, 8,611~records (52.78\% of the study dataset) received either a different active verb or additional component-level action detail. In 1,350~records (8.27\% of the study dataset), a proposed action populated an empty legacy field. Figure~\ref{fig:Figure_5} distinguishes retained active verbs with added detail from fully updated verbs.

Unexpected-action refinement initially covered 1,392~records with a specific failure-mode label and \texttt{Other} action. Clustering and record-level relabelling left 136~informative unmatched actions (0.83\% of the corpus), included in the accepted action-label total.

The dominant change from repair to replace follows the adopted definition: exchanging a subcomponent is replacement even when recorded historically as repair to its parent system. Component-level labels expose part consumption and the exchanged item, while retained legacy fields allow analysts to use either convention. Populating empty fields makes narrative-only interventions available categorically; adding component detail also enriches records whose active verb is unchanged.

% ..............................................................................
% SUBSUBSECTION 3.1.4: FAILURE-MODE EVIDENCE EXTRACTION
% ..............................................................................
\subsubsection{Failure-mode evidence extraction}
\label{subsubsec:failure_mode_results}

\begin{figure}[!ht]
    \centering
    \includegraphics[width=\linewidth]{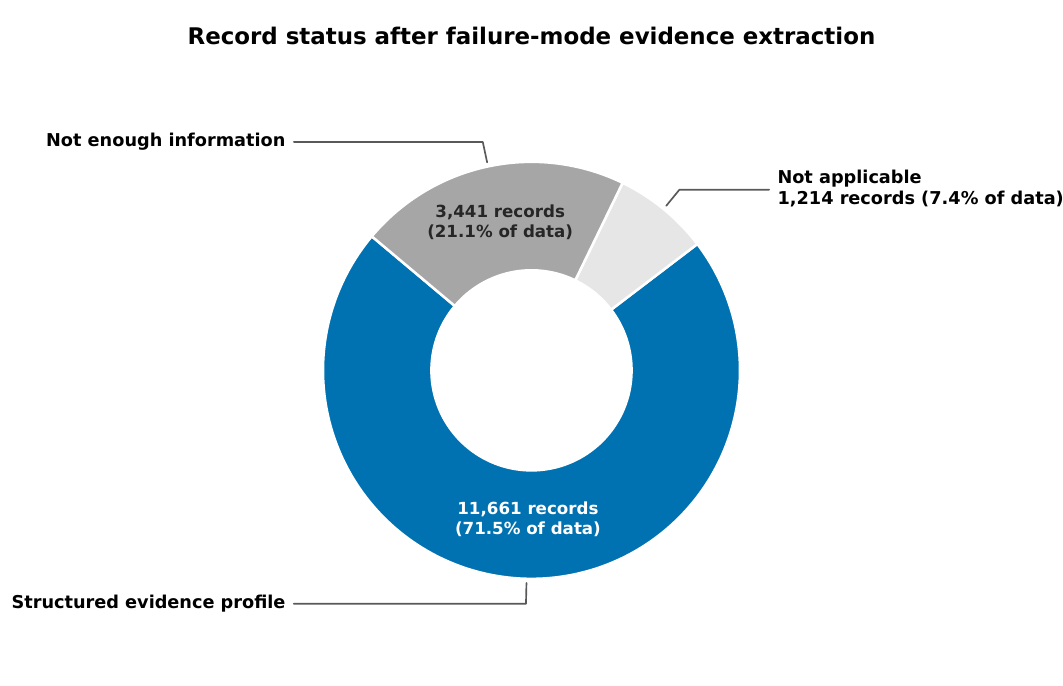}
    \caption[Failure-mode evidence extraction statuses]{Record status after failure-mode evidence extraction across all 16,316~maintenance records. Counts and percentages use the full dataset as the denominator. The accepted-label group includes 34~informative unmatched \texttt{Other} outcomes. The 1,214~\texttt{Not Applicable} statuses comprise 1,091~deterministic preventive/retrofit exclusions and 123~model-returned statuses.}
    \label{fig:Figure_6}
\end{figure}

Accepted failure-mode labels covered 11,661~records (71.47\%), including 34~informative unmatched \texttt{Other} outcomes (0.21\% of the corpus). Figure~\ref{fig:Figure_6} shows the full partition, with 3,441~unresolved records (21.09\%) and 1,214~inapplicable records (7.44\%).

The deterministic status rules accounted for 4,228~records: 3,137~received \texttt{Not enough information} because their actions were unresolved, and 1,091~received \texttt{Not Applicable}. The deterministic \texttt{Not Applicable} outcomes comprised 921~\emph{Preventive} and 170~\emph{Retrofit} records.

The remaining 12,088~records received 11,627~specific labels, 34~\texttt{Other} labels, 304~\texttt{Not enough information} outcomes, and 123~\texttt{Not Applicable} outcomes. Unresolved totals cover deterministic propagation and post-inference outcomes, including model-returned statuses and acceptance fallbacks. Inapplicable totals combine deterministic exclusions and model-returned statuses.

% ..............................................................................
% SUBSUBSECTION 3.1.5: UNRESOLVED LABELS AND EXCLUSION OUTCOMES
% ..............................................................................
\subsubsection{Unresolved labels and exclusion outcomes}
\label{subsubsec:unresolved_labels}

Across the 16,316~maintenance records, \texttt{Not enough information} was recorded for maintenance type in 2,065~records (12.66\%), action in 3,137 (19.23\%), and failure mode in 3,441 (21.09\%). All records with unresolved maintenance types also had unresolved actions and failure modes. A further 1,072~records had a maintenance-type label but unresolved action and failure-mode fields; 304~had labels available for both preceding fields but an unresolved failure mode. Available labels include \texttt{Other}.

Unspecified systems contribute the largest separately listed unresolved failure-mode count (604), followed by rotor blades (401). Table~\ref{tab:unresolved_labels} groups these outcomes by system and maintenance type, using full-corpus percentages rather than within-system unresolved rates.

An unspecified system does not necessarily preclude action or failure-mode labelling. For example, a description can support replacement of a fuse without identifying its system. A failure-mode label additionally requires evidence of the component's condition; replacement alone does not establish that condition.

\begin{table}[!htbp]
    \centering
    \footnotesize
    \setlength{\tabcolsep}{4pt}
    \renewcommand{\arraystretch}{1.1}
    \caption[Unresolved labels by system group and maintenance type]{Records with \texttt{Not enough information} outcomes by system and maintenance type. Cells give counts and percentages of all 16,316~records. Block A lists the seven groups with the largest unresolved failure-mode counts and combines the rest; block B includes all observed maintenance-type outcomes. Both include global totals.}
    \label{tab:unresolved_labels}
    \begin{tabularx}{\linewidth}{@{}>{\RaggedRight\arraybackslash}X >{\RaggedLeft\arraybackslash}p{3.0cm} >{\RaggedLeft\arraybackslash}p{2.6cm} >{\RaggedLeft\arraybackslash}p{2.6cm}@{}}
        \toprule
        \textbf{Category} & \textbf{Unresolved\newline maintenance type} & \textbf{Unresolved\newline action} & \textbf{Unresolved\newline failure mode} \\
        \midrule
        \multicolumn{4}{@{}l}{\textbf{A. By system group}} \\
        \addlinespace
        Not specified & 136 (0.83) & 587 (3.60) & 604 (3.70) \\
        Rotor Blades & 147 (0.90) & 341 (2.09) & 401 (2.46) \\
        Converter & 249 (1.53) & 260 (1.59) & 273 (1.67) \\
        Hub \& Main Shaft & 234 (1.43) & 250 (1.53) & 254 (1.56) \\
        Generator & 190 (1.16) & 226 (1.39) & 229 (1.40) \\
        Drive Train & 202 (1.24) & 227 (1.39) & 228 (1.40) \\
        Non-Pitch Hydraulics & 201 (1.23) & 218 (1.34) & 222 (1.36) \\
        Remaining Systems & 706 (4.33) & 1,028 (6.30) & 1,230 (7.54) \\
        \addlinespace
        \textbf{Total: all system groups} & \textbf{2,065 (12.66)} & \textbf{3,137 (19.23)} & \textbf{3,441 (21.09)} \\
        \midrule
        \multicolumn{4}{@{}l}{\textbf{B. By maintenance type}} \\
        \addlinespace
        Not enough information & & 2,065 (12.66) & 2,065 (12.66) \\
        Corrective & & 398 (2.44) & 499 (3.06) \\
        Preventive & & 465 (2.85) & 465 (2.85) \\
        Inspection & & 161 (0.99) & 336 (2.06) \\
        Predictive & & 42 (0.26) & 42 (0.26) \\
        Reset & & 6 (0.04) & 30 (0.18) \\
        Other & & 0 (0.00) & 4 (0.02) \\
        Retrofit & & 0 (0.00) & 0 (0.00) \\
        \addlinespace
        \textbf{Total: all maintenance types} & & \textbf{3,137 (19.23)} & \textbf{3,441 (21.09)} \\
        \bottomrule
    \end{tabularx}

    \vspace{2pt}
    \begin{minipage}{\linewidth}
        \footnotesize
        \textit{Note:} \emph{Remaining Systems} includes \emph{Pitch System} and every other unlisted group. Counts overlap across fields, and the two blocks cover the same dataset, so totals should not be added.
    \end{minipage}
\end{table}

The 465~preventive records with unresolved actions also received unresolved failure-mode outcomes because the action-status check preceded the preventive/retrofit exclusion (Section~\ref{subsubsec:acceptance_rules}).

Across all 3,441~records with unresolved failure modes, descriptions contained 3 to 97~words, with a mean of 22.20 and a median of 20. None were blank.

Qualitative review of the 2,065~records unresolved in all three fields identified recurring gaps in the maintenance histories. Descriptions included turbine-stop or alarm notifications, system headings, condition-monitoring updates, outstanding-work lists, and administrative closure. Additions such as ``work completed'', ``pending'', ``see details'', or a downtime-report reference often supplied no intervention narrative.

Maintenance-type interpretation required an identifiable intervention purpose; a recommendation could support \emph{Inspection} or another type even when completion was unconfirmed. Action interpretation required a clear activity and target, which a recommendation or plan could also provide. Alarm notifications without an identifiable intervention, generic instructions, and missing targets limited informative labelling. For failure modes, unexplained codes and threshold alarms could identify symptoms without a sufficiently specific component condition. Truncated text, missing referenced details, and inconsistent descriptions added ambiguity.

The review of all 1,214~\texttt{Not Applicable} records identified inspections and tests reporting no abnormality, scheduled lubrication and oil replenishment, software and equipment upgrades, and safety-compliance work such as replacement of expired supplies or age-limited equipment. Other descriptions concerned administrative tests, generic completion, or recommendations without confirmed findings. Such records often identify an action or maintenance purpose without documenting a failure condition.

The review also identified failure-related evidence in \texttt{Not Applicable} records describing failed aviation lights or communications, leakage, and corrosion. The sampled assessment quantifies approval of unresolved and inapplicable outcomes in Section~\ref{subsec:validation_results}.

% ------------------------------------------------------------------------------
% SUBSECTION 3.2: VALIDATION RESULTS
% ------------------------------------------------------------------------------
\subsection{Validation results}
\label{subsec:validation_results}

% ..............................................................................
% SUBSUBSECTION 3.2.1: ACHIEVED SUBSET BALANCE
% ..............................................................................
\subsubsection{Achieved subset balance}

The validation subset closely reproduced the four marginal outcome distributions used for selection (Table~\ref{tab:validation_subset}). The maximum absolute difference from an archive category share was 0.224~percentage points. Within the accepted system-code updates, the largest difference between correction-workflow shares was 0.285~percentage points, using 105~subset and 2,178~archive updates as the respective denominators.

\begin{table}[!htbp]
    \centering
    \footnotesize
    \setlength{\tabcolsep}{4pt}
    \renewcommand{\arraystretch}{1.1}
    \caption[Archive and validation-subset outcome distributions]{Comparison of the full archive and validation subset across the four balancing distributions. Entries are record counts (percentages), using 16,316~archive records and 800~validation records as the denominators in each block. Differences are subset minus archive in percentage points.}
    \label{tab:validation_subset}
    \begin{tabularx}{\linewidth}{@{}>{\RaggedRight\arraybackslash}X >{\RaggedLeft\arraybackslash}p{3.0cm} >{\RaggedLeft\arraybackslash}p{3.0cm} >{\RaggedLeft\arraybackslash}p{2.5cm}@{}}
        \toprule
        \textbf{Processing outcome or label} & \textbf{Full archive} & \textbf{Validation subset} & \textbf{Difference\newline (percentage points)} \\
        \midrule
        \multicolumn{4}{@{}l}{\textbf{System-code processing outcome}} \\
        \addlinespace
        Retained code or missing-code status & 14,138 (86.65\%) & 695 (86.88\%) & $+0.224$ \\
        Pitch correction & 1,095 (6.71\%) & 53 (6.62\%) & $-0.086$ \\
        Missing-code restoration & 496 (3.04\%) & 24 (3.00\%) & $-0.040$ \\
        Post-audit correction & 587 (3.60\%) & 28 (3.50\%) & $-0.098$ \\
        \midrule
        \multicolumn{4}{@{}l}{\textbf{Maintenance-type outcome}} \\
        \addlinespace
        Changed named category & 3,541 (21.70\%) & 174 (21.75\%) & $+0.047$ \\
        Retained named category & 10,254 (62.85\%) & 503 (62.88\%) & $+0.029$ \\
        Other & 456 (2.79\%) & 22 (2.75\%) & $-0.045$ \\
        Not enough information & 2,065 (12.66\%) & 101 (12.62\%) & $-0.031$ \\
        \midrule
        \multicolumn{4}{@{}l}{\textbf{Maintenance-action outcome}} \\
        \addlinespace
        Specific label & 13,043 (79.94\%) & 640 (80.00\%) & $+0.060$ \\
        Other & 136 (0.83\%) & 6 (0.75\%) & $-0.084$ \\
        Not enough information & 3,137 (19.23\%) & 154 (19.25\%) & $+0.023$ \\
        \midrule
        \multicolumn{4}{@{}l}{\textbf{Failure-mode outcome}} \\
        \addlinespace
        Specific label & 11,627 (71.26\%) & 570 (71.25\%) & $-0.011$ \\
        Other & 34 (0.21\%) & 2 (0.25\%) & $+0.042$ \\
        Not enough information & 3,441 (21.09\%) & 169 (21.13\%) & $+0.035$ \\
        Not Applicable & 1,214 (7.44\%) & 59 (7.37\%) & $-0.066$ \\
        \bottomrule
    \end{tabularx}

    \vspace{2pt}
    \begin{minipage}{\linewidth}
        \footnotesize
        \textit{Note:} Each block partitions the same archive and subset. Subset percentages use sum-preserving rounding to two decimal places so that each block totals 100.00\%. Differences were calculated from unrounded proportions. Changed and retained maintenance types refer to named categories; \texttt{Other} is separate regardless of its legacy value. Escape outcomes include model-returned statuses and those applied after failed acceptance or through deterministic processing.
    \end{minipage}
\end{table}

The 695~retained system outcomes comprised 654~specific codes and 41~unresolved outcomes, assessed separately in Table~\ref{tab:validation_subgroups}. All ten action-start years from 2017 to 2026 were represented.

% ..............................................................................
% SUBSUBSECTION 3.2.2: OVERALL APPROVAL
% ..............................................................................
\subsubsection{Overall approval}

\begin{figure}[!htbp]
    \centering
    \includegraphics[width=\linewidth]{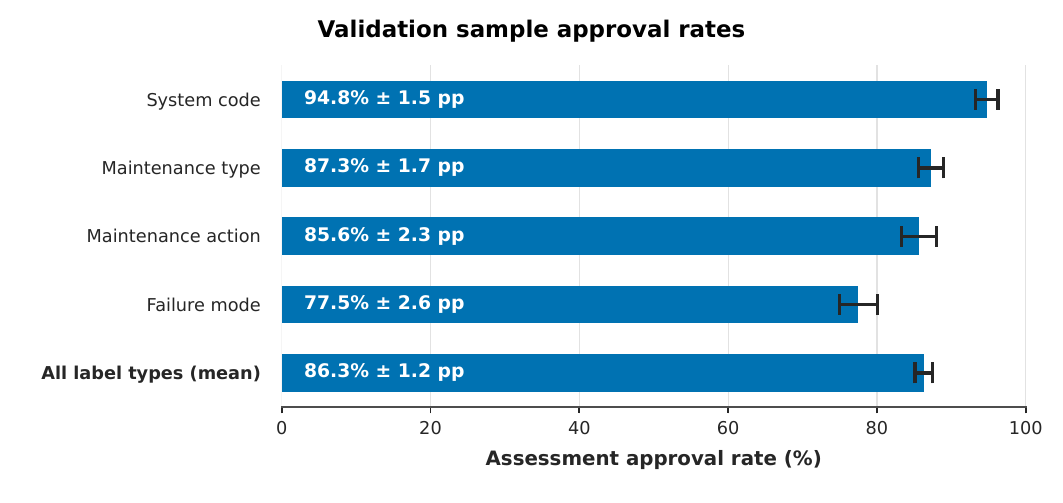}
    \caption[Field-level assessment approval]{Assessment approval for the four final fields in 800~maintenance records and their equal-weight mean. Failure-mode assessment covers titles and escape outcomes. Error bars show approximate 95\% sampling confidence intervals under the balanced-sampling variance approximation in Section~\ref{subsec:validation_design}; annotations give approval percentages and interval half-widths. The abbreviation pp denotes percentage points. Uncertainty for the mean uses records as the sampling units.}
    \label{fig:validation_overall_approval}
\end{figure}

System codes received approval in 758 of 800~records (94.8\%), maintenance types in 698 (87.3\%), actions in 685 (85.6\%), and failure-mode outcomes in 620 (77.5\%). Each denominator included specific labels and escape outcomes. Figure~\ref{fig:validation_overall_approval} shows these estimates with approximate 95\% confidence-interval half-widths of 1.5, 1.7, 2.3, and 2.6~percentage points, respectively.

The equal-weight cross-field mean was 86.3\%, representing 2,761~approved decisions out of 3,200, with an approximate interval half-width of 1.2~percentage points. Its uncertainty was estimated across the 800~sampled records using Equation~\ref{eq:validation_variance}.

% ..............................................................................
% SUBSUBSECTION 3.2.3: APPROVAL BY OUTCOME SUBGROUP
% ..............................................................................
\subsubsection{Approval by outcome subgroup}

System-code approval was similar across changed labels (94.3\%), unchanged specific labels (94.6\%), and unresolved outcomes (97.6\%; Table~\ref{tab:validation_subgroups}). Their judgement rules differ: changed codes required positive support, retained codes were approved unless contradicted, and unresolved outcomes required insufficient evidence for a code.

\begin{table}[!htbp]
    \centering
    \footnotesize
    \setlength{\tabcolsep}{4pt}
    \renewcommand{\arraystretch}{1.2}
    \caption[Assessment approval by outcome subgroup]{Approval by final-outcome subgroup in the 800-record validation subset. Cells show approval percentage (approved/assessed), rounded to one decimal place. The populated rows in each field partition the same 800~records. Failure-mode assessment covers titles and escape outcomes.}
    \label{tab:validation_subgroups}
    \begin{tabularx}{\linewidth}{@{}>{\RaggedRight\arraybackslash}X *{4}{>{\RaggedLeft\arraybackslash}p{3.0cm}}@{}}
        \toprule
        \textbf{Assessment subgroup} & \textbf{System code} & \textbf{Maintenance type} & \textbf{Maintenance action} & \textbf{Failure mode} \\
        \midrule
        Changed specific label & 94.3\% (99/105) & 64.9\% (113/174) & n.a. & n.a. \\
        Unchanged specific label & 94.6\% (619/654) & 99.2\% (499/503) & n.a. & n.a. \\
        Specific label & n.a. & n.a. & 84.7\% (542/640) & 87.7\% (500/570) \\
        Other & n.a. & 18.2\% (4/22) & 6/6 & 1/2 \\
        Unresolved & 97.6\% (40/41) & 81.2\% (82/101) & 89.0\% (137/154) & 46.7\% (79/169) \\
        Inapplicable & n.a. & n.a. & n.a. & 67.8\% (40/59) \\
        \bottomrule
    \end{tabularx}

    \vspace{2pt}
    \begin{minipage}{\linewidth}
        \footnotesize
        \textit{Note:} n.a. denotes a structurally inapplicable subgroup. Unresolved means \texttt{Not specified} for system codes and \texttt{Not enough information} for the other fields. Inapplicable means \texttt{Not Applicable}. The small action and failure-mode \texttt{Other} groups show counts only. Unchanged specific system codes and maintenance types use the retention criterion in Section~\ref{subsec:validation_design}; this default does not apply to escapes.
    \end{minipage}
\end{table}

Maintenance-type approval differed substantially between changed (64.9\%) and unchanged named labels (99.2\%). This contrast combines differences in outcome composition and assessment requirements. Approval of maintenance-type \texttt{Other} outcomes was particularly low (4/22, 18.2\%).

Specific action labels and failure-mode titles received approval in 84.7\% and 87.7\% of cases, respectively. Failure-mode escapes were less often approved: 46.7\% for unresolved and 67.8\% for inapplicable outcomes. Escapes accounted for 110 of the 180~failure-mode rejections (61.1\%), compared with 70~rejected specific titles. Overall failure-mode approval therefore reflects both title support and the appropriateness of withholding a title or declaring inapplicability.

% ..............................................................................
% SUBSUBSECTION 3.2.4: JOINT REJECTION PATTERNS
% ..............................................................................
\subsubsection{Joint rejection patterns}

\begin{figure}[!htbp]
    \centering
    \includegraphics[width=\linewidth]{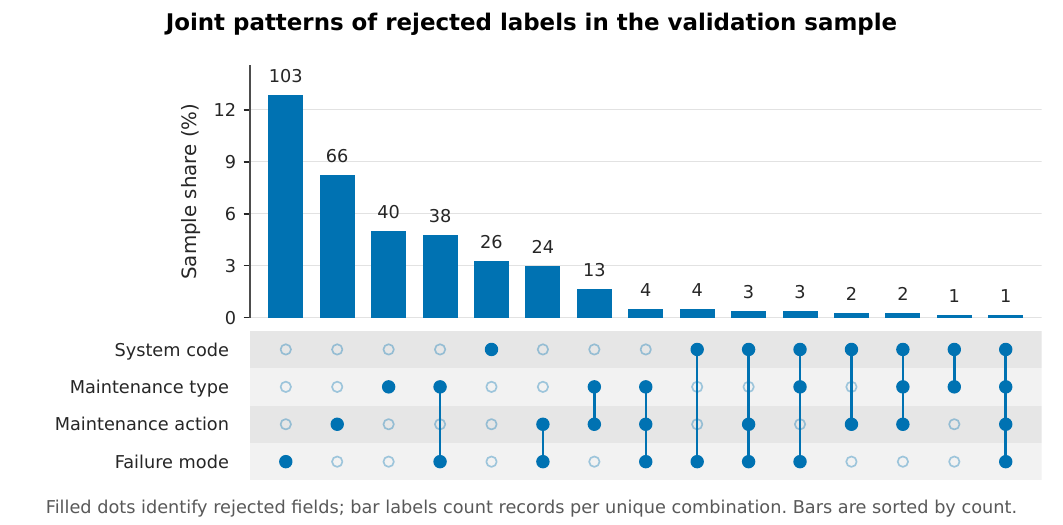}
    \caption[Joint patterns of assessment rejection]{Exact combinations of rejected fields in the 800-record validation subset. Filled dots identify rejected fields and hollow dots approved fields; each bar counts one mutually exclusive combination. Failure mode denotes assessment of titles and escape outcomes. Bar labels give record counts and bar heights express percentages of all 800~records. The 470~records with all fields approved are outside the displayed rejection combinations.}
    \label{fig:validation_joint_patterns}
\end{figure}

All four fields were approved in 470 of 800~records (58.8\%); the remaining 330~records had at least one rejected field. Figure~\ref{fig:validation_joint_patterns} partitions these rejections into 15~mutually exclusive combinations. The largest combination was rejection of the failure-mode field alone, affecting 103~records (12.9\% of the sample), followed by action-only rejection in 66~records. Maintenance-type and failure-mode outcomes were jointly rejected, with the other fields approved, in 38~records.

Exactly one field was rejected in 235~records, two in 82, three in 12, and all four in one. Multiple-field rejections affected 95~records (11.9\%). These patterns locate co-occurring assessment problems.

% ..............................................................................
% SUBSUBSECTION 3.2.5: QUALITATIVE ANALYSIS OF REJECTED OUTCOMES
% ..............................................................................
\subsubsection{Qualitative analysis of rejected outcomes}
\label{subsubsec:qualitative_rejections}

Qualitative review of the 439~rejected field decisions across 330 of the 800~validation records identified recurring mismatches between labels and the available evidence. System-code rejections included retained codes inconsistent with the affected functional system and proposed corrections more specific than the narrative supported.

Maintenance-type rejections concerned the distinction between an activity, such as cleaning or replacement, and the purpose of the intervention. In work order~\texttt{1131182}, correction of an excessive gearbox oil level occurred during annual maintenance, but the documented purpose remained remedial. Rejected \texttt{Other} outcomes included work that fitted a defined type and accounts with insufficient evidence of purpose. Rejected unresolved outcomes included records with identifiable corrective purposes or explicit inspection recommendations, even where inspection completion was unconfirmed.

Action-label rejections included composite descriptions containing unsupported activities or targets. Some records had unresolved action outcomes despite descriptions identifying a broad repair or inspection.

For specific failure-mode titles, rejection notes identified unsupported diagnostic detail. In one work order, an alarm and improvement after replacement supported a generator-bearing defect without establishing rolling-contact fatigue. Unresolved failure-mode outcomes included records supporting informative alarm conditions or broad abnormalities, as well as records warranting a non-failure interpretation. Rejected \texttt{Not Applicable} outcomes overlooked abnormal conditions in accounts that also described servicing or modification.

The shared issue was alignment between label specificity and the supporting account. Unsupported detail misrepresented the evidence, while inappropriate escape outcomes discarded informative content. A broad description could be supported even when a finer component, intervention or diagnostic distinction remained unresolved.

% ------------------------------------------------------------------------------
% SUBSECTION 3.3: PROJECT ECONOMICS AND COMPUTATIONAL LOAD
% ------------------------------------------------------------------------------
\subsection{Project economics and computational load}
\label{sec:project_economics}

% ..............................................................................
% SUBSUBSECTION 3.3.1: FINANCIAL EXPENDITURE, RUNTIME, AND THROUGHPUT
% ..............................................................................
\subsubsection{Financial expenditure, runtime, and throughput}

\begin{figure}[!ht]
    \centering
    \includegraphics[width=\linewidth]{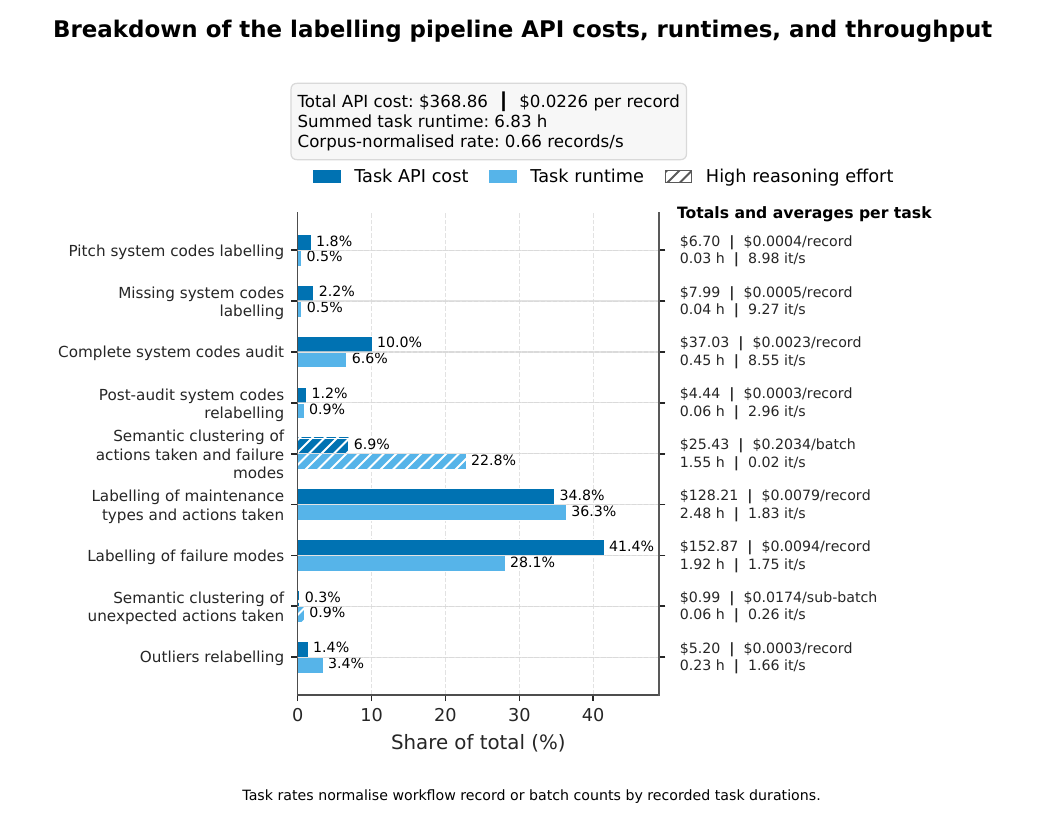}
    \caption[API expenditure, runtime, and throughput by task]{Recorded application programming interface (API) expenditure and runtime, with workflow-normalised throughput by task. Bars show shares of total expenditure and summed task runtime; right-hand annotations give task totals and averages. Per-record costs divide task expenditure by the full 16,316-record corpus; batch and sub-batch costs use the respective task counts. The summary rate is corpus size divided by the summed recorded task duration. Task rates use each task's workflow record, batch or sub-batch count divided by its recorded duration; \texttt{it/s} denotes items per second.}
    \label{fig:Figure_9}
\end{figure}

Recorded API expenditure was \$368.86 at execution-time prices, or \$0.0226~per corpus record. Recorded task durations totalled 6.83~hours under the configured asynchronous limits (Figure~\ref{fig:Figure_9}).

Figure~\ref{fig:master_workflow} groups the pitch and missing-code tasks as targeted system-code classification and represents semantic clustering of actions and failure modes as batch taxonomy extraction. Figures~\ref{fig:master_workflow} and~\ref{fig:Figure_9} distinguish unexpected-action clustering from the subsequent record-level action-relabelling pass.

The corpus-normalised processing rate was 0.66~records per second, calculated by dividing the 16,316-record corpus by the summed recorded task duration of 24,598~s. Task rates use the corresponding workflow record, batch or sub-batch counts. Batch extraction used larger contexts, larger dictionaries and higher reasoning effort than record-level system-code tasks. Runtime reflects these differences alongside concurrency limits and provider-side behaviour.

% ..............................................................................
% SUBSUBSECTION 3.3.2: TOKEN UTILISATION
% ..............................................................................
\subsubsection{Token utilisation}

\begin{figure}[!ht]
    \centering
    \includegraphics[width=\linewidth]{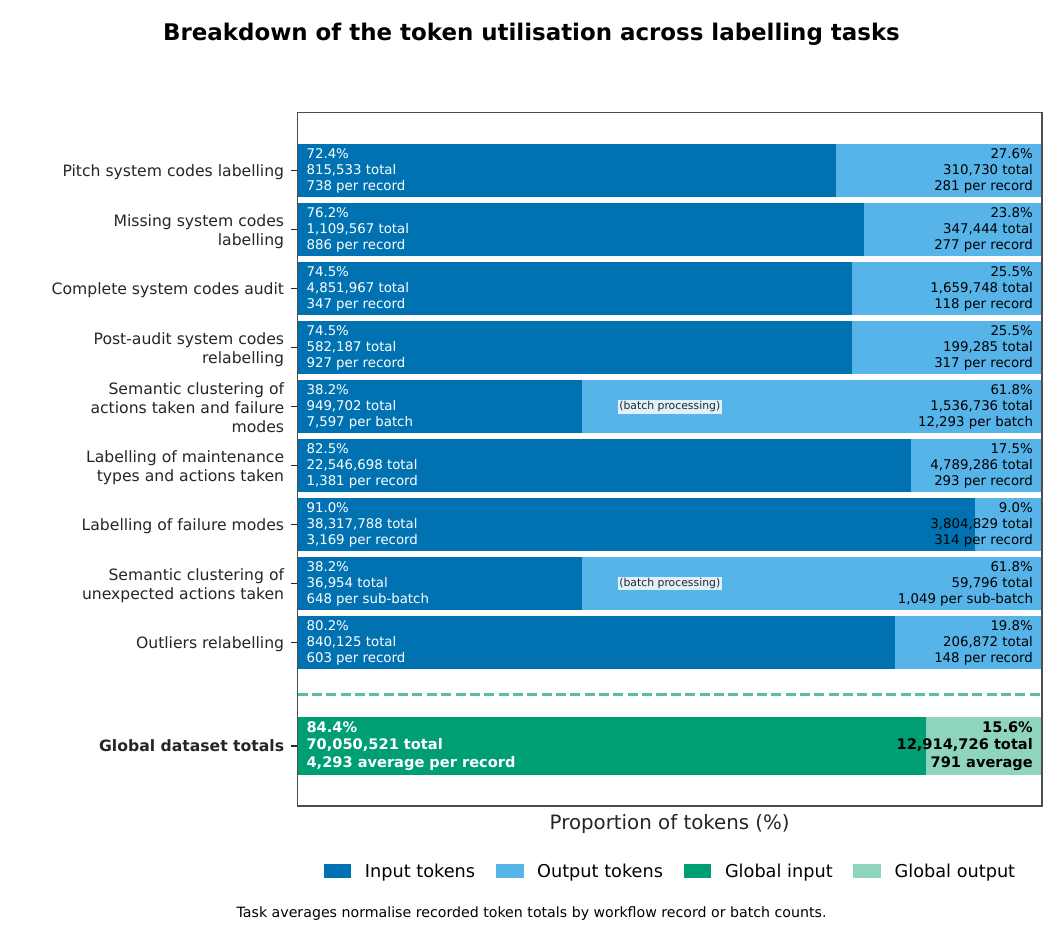}
    \caption[Token utilisation by pipeline task]{Recorded input and output token shares, totals, and normalised averages. Tokens are text units processed or generated by the model. Task percentages use combined input and output tokens within each row; global percentages use all recorded tokens. Task averages are recorded token totals divided by the corresponding workflow record, batch or sub-batch count. Global per-record averages use the full 16,316-record corpus.}
    \label{fig:Figure_10}
\end{figure}

The workload comprised 70,050,521~input tokens (84.4\%) and 12,914,726~output tokens (15.6\%), where tokens are units of text processed or generated by the model (Figure~\ref{fig:Figure_10}). Record-level tasks consumed more input than output because they repeatedly supplied taxonomies and narratives for compact responses. Batch extraction reversed this balance (61.8\% output, 38.2\% input), generating detailed candidate dictionaries. Repeated static context and dictionary length are therefore distinct targets for prompt and batch design.

% ------------------------------------------------------------------------------
% SUBSECTION 3.4: EXAMPLES OF DESCRIPTIVE OUTPUTS ENABLED BY SEMANTIC ENRICHMENT
% ------------------------------------------------------------------------------
\subsection{Examples of descriptive outputs enabled by semantic enrichment}
\label{subsec:descriptive_outputs}

Figure~\ref{fig:Figure_11} compares a legacy turbine-record baseline with an enriched population retained by the candidate flag (Section~\ref{subsubsec:candidate_record_selection}). Figures~\ref{fig:Figure_12} and~\ref{fig:Figure_13} summarise the flag-selected turbine records using the workflow outputs. The candidate flag does not distinguish planned from performed activities. Informative actions can support inclusion despite an unresolved or inapplicable failure mode. The nested counts in Table~\ref{tab:unresolved_labels} imply that 125~records with unresolved failure modes remain eligible across the full corpus before restriction to turbine records: 101~corrective and 24~reset records. These descriptive results should be interpreted alongside the assessment in Section~\ref{subsec:validation_results}.

% ..............................................................................
% SUBSUBSECTION 3.4.1: CHANGES IN CANDIDATE FUNCTIONAL-FAILURE RECORD SHARES
% ..............................................................................
\subsubsection{Changes in candidate functional-failure record shares}
\label{subsubsec:system_share_changes}

\begin{figure}[!ht]
    \centering
    \includegraphics[width=\linewidth]{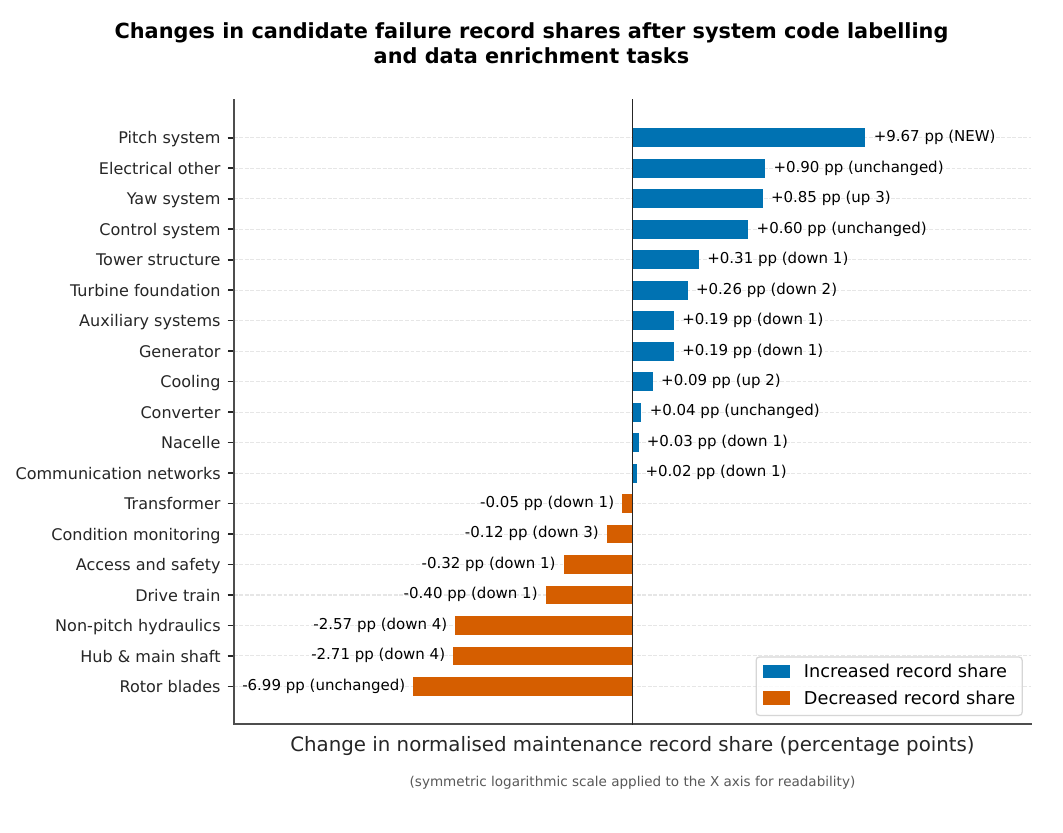}
    \caption[Changes in candidate functional-failure record shares]{Changes in turbine system shares between 13,758~legacy records and 10,649~enriched records retained by the candidate flag (Section~\ref{subsubsec:candidate_record_selection}). Both populations exclude unspecified systems; only the enriched population is filtered by the flag. Each population is separately normalised to 100\%; values are percentage-point differences between system shares. Parentheses show rank changes; \textit{new} denotes a separated category. The symmetric logarithmic horizontal scale is linear within $\pm 0.5$~percentage points and compresses larger magnitudes to keep smaller changes visible; numerical annotations give exact differences.}
    \label{fig:Figure_11}
\end{figure}

Figure~\ref{fig:Figure_11} illustrates how classification and candidate selection jointly change relative system-share patterns between the legacy and enriched turbine-record populations. The legacy baseline contains 13,758~records after excluding 1,251~with unspecified systems from 15,009~turbine records. The enriched population contains 10,649~records after excluding 123~with unspecified systems from 10,772~flag-selected turbine records.

The newly separated pitch category shows the largest increase in candidate-record share (9.67~percentage points; Figure~\ref{fig:Figure_11}). Shares fall by 6.99~percentage points for rotor blades, 2.71 for hub and main shaft, and 2.57 for non-pitch hydraulics. Each other displayed system changes by at most 0.90~percentage points. The combined effects of taxonomy revision and filtering are therefore concentrated in particular systems.

The redistribution combines classification and filtering, rather than isolating the exclusion of unresolved or inapplicable outcomes. Separate normalisation means that a system's share can also change when records elsewhere are removed or reassigned.

% ..............................................................................
% SUBSUBSECTION 3.4.2: ASSIGNED MAINTENANCE-ACTION LABEL DISTRIBUTIONS
% ..............................................................................
\subsubsection{Assigned maintenance-action label distributions}
\label{subsubsec:action_distributions}

\begin{figure}[!ht]
    \centering
    \includegraphics[width=\linewidth]{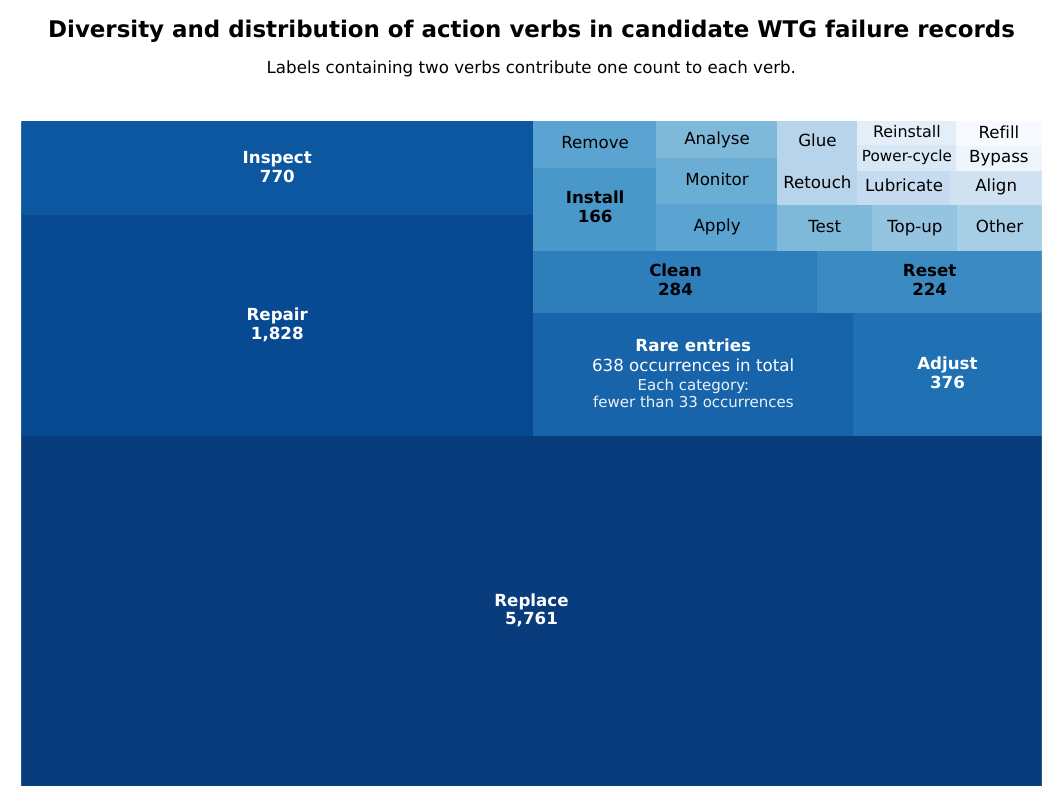}
    \caption[Distribution of maintenance-action verbs in selected turbine records]{Distribution of 10,938~occurrences of action verbs and \texttt{Other}, extracted from 10,772~turbine maintenance records retained by the candidate-record flag (Section~\ref{subsubsec:candidate_record_selection}). Areas encode occurrence counts; labels containing multiple verbs contribute one count to each verb. The generic \texttt{Other} action category denotes informative unmatched outcomes. The rare-entry group contains 638~occurrences, with fewer than 33~in each category. WTG denotes wind turbine generator.}
    \label{fig:Figure_12}
\end{figure}

\begin{figure}[!ht]
    \centering
    \includegraphics[width=\linewidth]{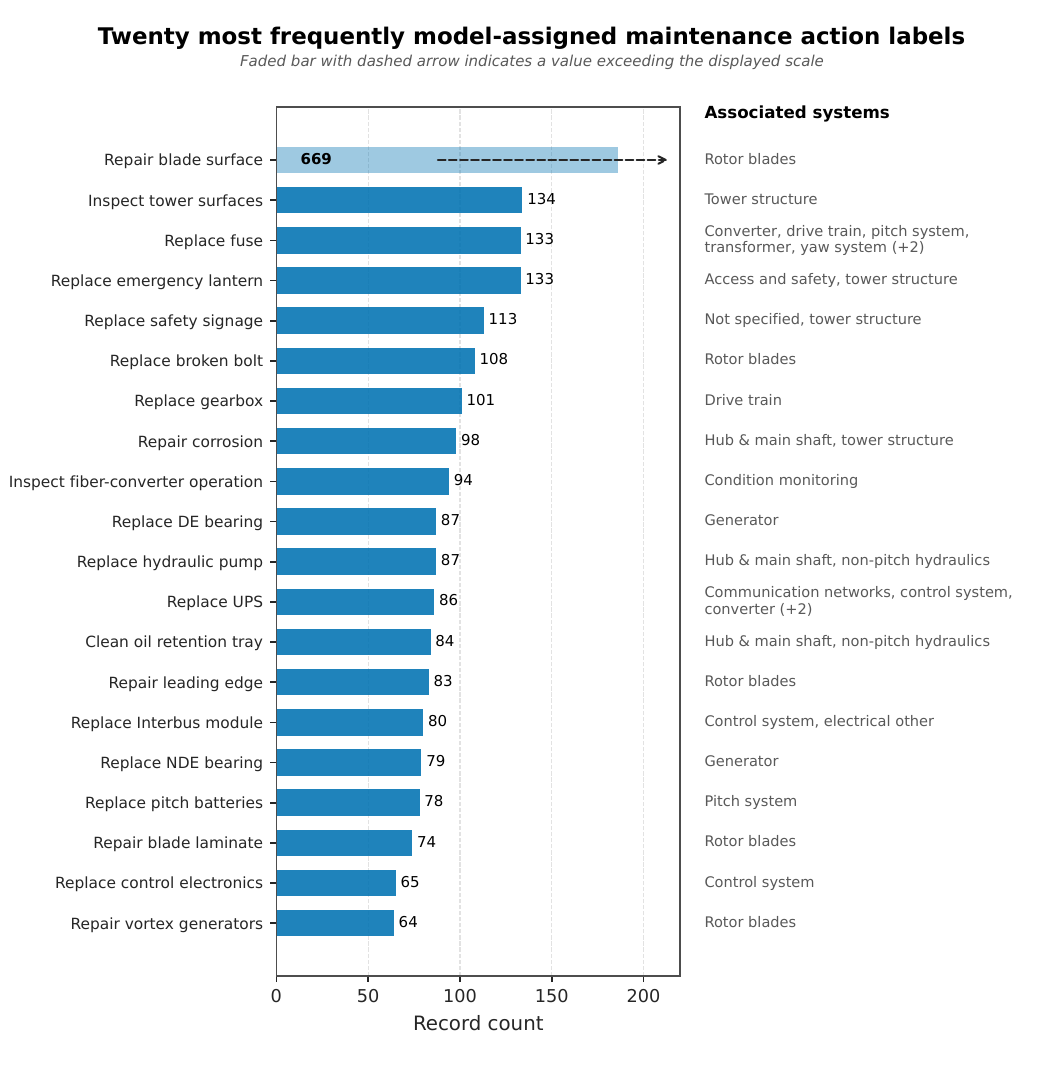}
    \caption[Twenty most frequent accepted maintenance-action labels]{The twenty most frequent model-assigned maintenance-action labels among turbine records retained by the candidate-record flag (Section~\ref{subsubsec:candidate_record_selection}), excluding generic \texttt{Other} labels. Bar lengths encode maintenance-record counts, with associated turbine systems shown at right. The faded leading bar and dashed extension indicate a value exceeding the common display scale. DE denotes drive end, NDE non-drive end, and UPS uninterruptible power supply.}
    \label{fig:Figure_13}
\end{figure}

Replacement (5,761~occurrences) and repair (1,828) dominate the 10,938~verb/category occurrences from 10,772~selected turbine records in Figure~\ref{fig:Figure_12}. Inspection, adjustment, cleaning, reset, and installation expose further activities obscured by generic legacy actions. Figure~\ref{fig:Figure_13} instead counts records by full action label, excluding \texttt{Other}. All 3,137~records with unresolved actions in the full corpus also had unresolved failure modes and failed candidate selection.

The leading displayed action label, \emph{Repair blade surface}, covers 669~records. Labels for leading-edge, laminate, bolt, and accessory work identify further rotor elements, while emergency-lantern and signage activity concerns safety provisions. Fuse replacement occurs across seven functional systems, and fan replacement also occurs in different systems. Retaining the action and associated system distinguishes these workloads beyond their shared verbs.

A targeted qualitative review of source descriptors supported this interpretation, separately from the 800-record assessment in Section~\ref{subsec:validation_results}. Repeated instructions to check condition-monitoring fibre converters did not establish converter failure. Cleaning oil-retention trays addressed a leakage consequence without necessarily correcting its source. The detailed labels therefore help select diagnostic, safety, housekeeping, and corrective work for engineering review.

Personal consultations with the operator that supplied the data helped explain the blade workload. The operator described much recurrent blade work as minor maintenance intended to sustain service life. Whole-blade replacement was described as costly, with replacement blades difficult to procure for older turbine models, motivating more frequent maintenance activities. The operator also reported an aerodynamic-upgrade campaign that increased the frequency of associated work orders during that period. Such upgrades are distinct from repair following accessory detachment. The consultations provide operational context for the observed maintenance patterns. Section~\ref{subsubsec:gearbox_action_review} examines source support for \emph{Replace gearbox}.

% ..............................................................................
% SUBSUBSECTION 3.4.3: FREQUENT FAILURE-MODE EVIDENCE LABELS
% ..............................................................................
\subsubsection{Frequent failure-mode evidence labels}
\label{subsubsec:frequent_failure_modes}

Table~\ref{tab:frequent_failure_modes} presents ten accepted failure-mode labels selected from frequently assigned categories across several turbine systems. Their profiles describe common mechanisms, possible symptoms, and candidate causes, extending beyond the title-level assessment in Section~\ref{subsec:validation_results}. This organisation provides a starting taxonomy for subsequent FMEA development~\cite{scheu_systematic_2019,dinmohammadi_fuzzy-fmea_2013}.

Blade shell or coating damage is the largest displayed category (704~records). Other profiles distinguish generator bearings from monitoring sensors and fibre converters, and InterBus hardware from power-converter boards. These distinctions permit component-specific case selection beyond broad system headings. Some profiles in the full output were assigned to only one record.

\begin{sidewaystable}[p]
    \centering
    \scriptsize
    \setlength{\tabcolsep}{4pt}
    \renewcommand{\arraystretch}{1.1}

    \caption[Selected frequent failure-mode evidence labels]{Ten accepted, model-assigned failure-mode labels selected from frequently occurring categories across turbine systems. Counts refer to maintenance records within the 11,661-record accepted-label corpus.}
    \label{tab:frequent_failure_modes}

    \newcolumntype{Y}[1]{>{\hsize=#1\hsize\linewidth=\hsize\RaggedRight\arraybackslash}X}

    \begin{tabularx}{\linewidth}{@{} c Y{0.90} Y{0.95} Y{1.10} Y{1.05} >{\RaggedLeft\arraybackslash}p{1.0cm} >{\RaggedRight\arraybackslash}p{2.0cm} @{}}
    \toprule
    \textbf{No.} & \textbf{Failure Mode} & \textbf{Common Mechanisms} & \textbf{Possible Symptoms} & \textbf{Candidate Causes} & \textbf{Record\newline Count} & \textbf{Turbine System} \\
    \midrule

    1 & Blade shell or coating damage requiring repair & Surface coating breakdown, laminate wear or local delamination of the composite shell & Categories 3 and 4 damage reports, visible blade defects during inspections, and general blade repair campaigns & Long-term environmental exposure, cyclic loading, minor impacts, or inspection-identified damage with exact cause not recorded & 704 & Rotor Blades \\ \cmidrule[0.2pt](lr){1-7}

    2 & Blade root attachment bolt fracture or loosening & Fatigue fracture, preload loss, and load redistribution in the blade-to-hub fastener set & Broken bolts/screws at the blade root, pitch-related alarms, extraction of failed bolts, and replacement of adjacent bolts or full retrofit sets & High cyclic pitch loads, fretting/corrosion, insufficient retained preload, and stress concentration at individual bolts & 127 & Rotor Blades \\ \cmidrule[0.2pt](lr){1-7}

    3 & Drive-end bearing rolling-contact fatigue & Raceway or rolling-element fatigue produces spalling, increased vibration and progressive mechanical noise & Condition monitoring system (CMS) vibration growth on the drive-end side, abnormal noise, urgent bearing recommendations and drive-end bearing replacement & Cyclic radial/axial loading, vibration stress, contamination and service-age fatigue & 122 & Generator \\ \cmidrule[0.2pt](lr){1-7}

    4 & InterBus node, input/output card or bus controller failed & InterBus module, card, head station or associated bus electronics lost function & \texttt{FM700}, \texttt{FM300}, \texttt{FM345} or related bus stop faults, communication loss and turbine stoppage & Electronic ageing, vibration, moisture ingress or thermal stress on input/output hardware & 116 & Control System \\ \cmidrule[0.2pt](lr){1-7}

    5 & Converter control, distribution, or driver printed circuit board fails & Electronic component burnout, logic failure, or signal routing loss on control and thread boards & Class A/B/C faults, thread shutdowns, control loss, or replacement of \texttt{TDB}, \texttt{AEAB}, \texttt{AEBI}, \texttt{AEPS}, \texttt{AEDB}, \texttt{DB}, \texttt{CCB}-class boards & Thermal ageing, over-voltage stress, contamination, and collateral damage from upstream power faults & 114 & Converter \\ \cmidrule[0.2pt](lr){1-7}

    6 & Vibration sensor becomes detached or shifts position & Adhesive bond or sensor base loses mechanical integrity and the sensor moves on the mounting surface & Channel goes offline, vibration values drift, or the sensor must be re-glued/repositioned & Inadequate bonding, incorrect adhesive or base preparation, maintenance disturbance, or vibration loading & 111 & Condition Monitoring \\ \cmidrule[0.2pt](lr){1-7}

    7 & Non-drive-end bearing rolling-contact fatigue & Raceway or rolling-element fatigue produces spalling, envelope growth and progressive vibration increase & CMS alarms on the non-drive-end side, abnormal noise, severity escalation and non-drive-end bearing replacement & Cyclic loading, vibration stress, contamination and service-age fatigue on the non-drive-end side & 110 & Generator \\ \cmidrule[0.2pt](lr){1-7}

    8 & CMS communication path fails at fibre converter & Electro-optical conversion or converter-side power or input arrangement fails & No remote communication until the base, top, or nacelle converter is replaced or re-powered & Fibre-converter hardware degradation or unsuitable converter power-supply arrangement & 109 & Condition Monitoring \\ \cmidrule[0.2pt](lr){1-7}

    9 & Hydraulic oil leakage into hub retention tray & Oil escapes from the pitch or hub hydraulic circuit and accumulates in the tray or leakage tank & Oil leakage alarm or errors, oil found in the hub or tray, repeated cleaning, drainage, and occasional oil level correction & Minor seal seepage, loosened joints, small hose leaks, or overflow sources not isolated in the log & 89 & Hub \& Main Shaft \\ \cmidrule[0.2pt](lr){1-7}

    10 & Main inverter stator/line/control contactor or relay fails & Contact welding, coil burnout, auxiliary feedback loss, or relay dropout in \texttt{MI} switching circuits & \texttt{FM300}, \texttt{FM1201}, \texttt{FM1208}, or \texttt{FM1209} trips, no-ready states, or repeated contactor/relay replacement & Electrical arcing, thermal ageing, loose neutral/auxiliary wiring, and high switching duty & 86 & Converter \\

    \bottomrule
    \end{tabularx}

    \vspace{2pt}
    \begin{minipage}{\linewidth}
        \scriptsize
        \textit{Note:} Common mechanisms summarise representative engineering context at profile level. CMS denotes condition monitoring system. Fault codes \texttt{FM700}, \texttt{FM300}, \texttt{FM345}, \texttt{FM1201}, \texttt{FM1208}, and \texttt{FM1209}, and identifiers \texttt{TDB}, \texttt{AEAB}, \texttt{AEBI}, \texttt{AEPS}, \texttt{AEDB}, \texttt{DB}, \texttt{CCB}, and \texttt{MI}, retain source terminology. Categories 3 and 4 are legacy blade-damage descriptors whose grading criteria are unavailable. Converter threads denote numbered branches in the source records.
    \end{minipage}
\end{sidewaystable}

% ..............................................................................
% SUBSUBSECTION 3.4.4: TOPOLOGY-AWARE PITCH-ARCHITECTURE COMPARISON
% ..............................................................................
\subsubsection{Topology-aware pitch-architecture comparison}
\label{subsubsec:pitch_comparison}

Table~\ref{tab:pitch_modes_comparison} contrasts electrical and hydraulic pitch labels using maintenance-record counts. Electrical-pitch labels centre on emergency power, drives, and control, led by battery capacity loss (65~records). Hydraulic labels centre on leakage, mechanical play, pressure storage, cylinders, pumps, and valves, led by combined leakage and play (29~records). Controller area network (CAN) communication faults identify an electrical-control interface within the hydraulic architecture. The dictionaries thus distinguish actuation, energy storage, and control within each topology.

\begin{table}[!htbp]
    \centering
    \scriptsize
    \setlength{\tabcolsep}{4pt}
    \renewcommand{\arraystretch}{1.1}
    \caption[Failure-mode evidence labels by pitch architecture]{Seven most frequent model-assigned failure-mode labels within the electrical and hydraulic pitch-system groups, comprising 130 and 150~turbines, respectively. Values are maintenance-record counts. CAN denotes controller area network.}
    \label{tab:pitch_modes_comparison}

    \begin{tabularx}{\linewidth}{@{} c >{\RaggedRight\arraybackslash}X >{\RaggedRight\arraybackslash}X @{}}
    \toprule
    \textbf{No.} & \textbf{Electrical Pitch Failure Modes and Candidate Causes, $n=130$~Turbines} & \textbf{Hydraulic Pitch Failure Modes and Candidate Causes, $n=150$~Turbines} \\
    \midrule

    1 &
    \textbf{Mode:} Pitch battery capacity loss or voltage collapse\newline
    \textbf{Candidate causes:} Calendar ageing, repeated charge-discharge cycling, prolonged float service, and insufficient recharge margin\newline
    \textbf{Subsystem:} Pitch Emergency Power Supply/Batteries\newline
    \textbf{Frequency:} 65 records &
    \textbf{Mode:} Combined hydraulic leakage and mechanical play in the pitch system\newline
    \textbf{Candidate causes:} Cumulative age-related wear under cyclic loading and repeated hydraulic pressurisation\newline
    \textbf{Subsystem:} Pitch System (General)\newline
    \textbf{Frequency:} 29 records \\
    \cmidrule[0.2pt](lr){1-3}

    2 &
    \textbf{Mode:} Battery contactor fails to switch or provide stable feedback\newline
    \textbf{Candidate causes:} High switching duty, load arcing, component ageing, and vibration exposure\newline
    \textbf{Subsystem:} Pitch Emergency Power Supply/Batteries\newline
    \textbf{Frequency:} 30 records &
    \textbf{Mode:} Controller area network (CAN) communication cable or connection fault\newline
    \textbf{Candidate causes:} Vibration-loosened terminations, conductor fatigue, and connector oxidation in CAN wiring\newline
    \textbf{Subsystem:} Pitch System (General)\newline
    \textbf{Frequency:} 20 records \\
    \cmidrule[0.2pt](lr){1-3}

    3 &
    \textbf{Mode:} Pitch inverter internal electronics failure\newline
    \textbf{Candidate causes:} Ageing, thermal cycling, and electrical stress of inverter electronics\newline
    \textbf{Subsystem:} Pitch Motor Converter/Inverter\newline
    \textbf{Frequency:} 26 records &
    \textbf{Mode:} Accumulator pressure storage capacity degraded\newline
    \textbf{Candidate causes:} Age-related gas permeation, elastomer fatigue, and repeated pressure-cycling wear inside the accumulator\newline
    \textbf{Subsystem:} Pitch Hydraulic Accumulator\newline
    \textbf{Frequency:} 17 records \\
    \cmidrule[0.2pt](lr){1-3}

    4 &
    \textbf{Mode:} Battery charger fails to charge or regulate the battery string\newline
    \textbf{Candidate causes:} Thermal ageing of charger electronics, long service exposure, and electrical overstress\newline
    \textbf{Subsystem:} Pitch Emergency Power Supply/Batteries\newline
    \textbf{Frequency:} 23 records &
    \textbf{Mode:} Pitch cylinder/actuator functional failure\newline
    \textbf{Candidate causes:} Wear, contamination, or fatigue damage not further specified in the logs\newline
    \textbf{Subsystem:} Pitch Pressure Cylinder\newline
    \textbf{Frequency:} 13 records \\
    \cmidrule[0.2pt](lr){1-3}

    5 &
    \textbf{Mode:} Pitch drive motor internal electrical or mechanical failure\newline
    \textbf{Candidate causes:} Accumulated thermal ageing, repetitive pitch duty, vibration, and wear of internal motor components\newline
    \textbf{Subsystem:} Pitch Drive Motor\newline
    \textbf{Frequency:} 23 records &
    \textbf{Mode:} Pitch cylinder external oil leakage\newline
    \textbf{Candidate causes:} Seal ageing, pressure cycling, and contamination-induced wear\newline
    \textbf{Subsystem:} Pitch Pressure Cylinder\newline
    \textbf{Frequency:} 13 records \\
    \cmidrule[0.2pt](lr){1-3}

    6 &
    \textbf{Mode:} Pitch master electronic module failure\newline
    \textbf{Candidate causes:} Thermal cycling, vibration, and age-related electronic component wear\newline
    \textbf{Subsystem:} Pitch Controller\newline
    \textbf{Frequency:} 21 records &
    \textbf{Mode:} Hydraulic pump cannot maintain required pitch pressure\newline
    \textbf{Candidate causes:} Progressive abrasion, fatigue wear, or seal degradation of pump components\newline
    \textbf{Subsystem:} Pitch Hydraulic Oil Pump\newline
    \textbf{Frequency:} 10 records \\
    \cmidrule[0.2pt](lr){1-3}

    7 &
    \textbf{Mode:} Pitch brake assembly fails to release or hold as intended\newline
    \textbf{Candidate causes:} Ageing, repeated duty-cycle thermal cycling, and service wear; exact initiating subcomponent not recorded in the logs\newline
    \textbf{Subsystem:} Pitch Drive\newline
    \textbf{Frequency:} 20 records &
    \textbf{Mode:} Proportional or pitch control valve failure\newline
    \textbf{Candidate causes:} Hydraulic contamination, coil degradation, and thermal ageing of electro-hydraulic valve components\newline
    \textbf{Subsystem:} Pitch System (General)\newline
    \textbf{Frequency:} 10 records \\

    \bottomrule
    \end{tabularx}
\end{table}

These component families have a qualitative counterpart in field reliability evidence. Walgern et al.~\cite{walgern_reliability_2023} identify battery packs and grouped control and relay components in electrical pitch systems, and accumulators, cylinders, and valves in hydraulic systems. The present labels recover corresponding energy-storage, control, and actuation distinctions from maintenance narratives. This correspondence supports the engineering relevance of the categories.

% ------------------------------------------------------------------------------
% SUBSECTION 3.5: QUALITATIVE REVIEW OF SEMANTIC LABELS
% ------------------------------------------------------------------------------
\subsection{Qualitative review of semantic labels}
\label{subsec:semantic_label_review}

Two selected post-labelling reviews complement the sampled assessment (Section~\ref{subsec:validation_results}) and archive-wide source readings (Section~\ref{subsubsec:unresolved_labels}): gearbox action support and failure-mode category coherence.

% ..............................................................................
% SUBSUBSECTION 3.5.1: GEARBOX REPLACEMENT ACTION REVIEW
% ..............................................................................
\subsubsection{Gearbox replacement action review}
\label{subsubsec:gearbox_action_review}

The \emph{Replace gearbox} label was assigned to 101~work orders. This count appeared high relative to the operator's experience and prior knowledge of approximate replacement numbers, prompting a targeted review of the selected orders' descriptions, later notes, dates, and references to other orders on the same turbine. Table~\ref{tab:gearbox_action_review} separates reported work from repeated accounts, plans, and related activity.

\begin{table}[!htbp]
    \centering
    \footnotesize
    \setlength{\tabcolsep}{4pt}
    \renewcommand{\arraystretch}{1.1}
    \caption[Selected review of gearbox action labels]{Narrative review of the 101~work orders labelled \emph{Replace gearbox}. Categories group reported work, repeated accounts, plans, and related activities.}
    \label{tab:gearbox_action_review}
    \begin{tabularx}{\linewidth}{@{}X r@{}}
        \toprule
        \textbf{Narrative review category} & \textbf{Work orders} \\
        \midrule
        Main account of a reported replacement, anchoring a provisional group & 48 \\
        Further or potentially duplicate account & 34 \\
        Replacement planned or suggested, completion unconfirmed & 9 \\
        Administrative or related work & 4 \\
        Unresolved order & 6 \\
        \midrule
        \textbf{Total selected work orders} & \textbf{101} \\
        \bottomrule
    \end{tabularx}
\end{table}

The 48~main accounts define provisional narrative groups for reviewing reported replacements. Nine orders described planned or suggested replacements without confirming completion. These descriptions can support the \emph{Replace gearbox} action label.

The six unresolved orders mentioned gearbox or component replacement in residual alarm text or otherwise uninformative descriptions. Interpreting these mentions required additional information. An order concerning a damaged electrical box referred to a separate replacement; this reference alone did not support \emph{Replace gearbox} for that order. Multiple accounts of one reported replacement primarily raise a counting problem. Assessing label support requires interpreting each order's own account.

An independent asset register was unavailable, and the audit extract lacked model rationales and acceptance metadata. The review identifies source-support questions and potentially repeated documentation for subsequent expert follow-up.

% ..............................................................................
% SUBSUBSECTION 3.5.2: ENGINEERING PLAUSIBILITY AND COHERENCE OF FAILURE-MODE PROFILES
% ..............................................................................
\subsubsection{Engineering plausibility and coherence of failure-mode profiles}
\label{subsec:profile_coherence_review}

The qualitative review assessed the engineering plausibility of the displayed profiles, considering titles, parent-system labels, common mechanisms, possible symptoms, and candidate causes. It examined selected cases in greater depth, targeting intervention or alarm descriptions, overlapping categories, ambiguous system placement, and mixed functional outcomes or field meanings. This profile-level appraisal complements the record-level title assessment in Section~\ref{subsec:validation_results}.

The profiles in Table~\ref{tab:frequent_failure_modes} contain recognisable relationships between component condition, common mechanisms, and possible symptoms relevant to onshore turbine maintenance. The sensor-attachment profile connects loss of bond or base integrity with movement and altered signal behaviour; re-bonding describes the maintenance response. The blade-root fastener profile links preload loss, load redistribution and fatigue with descriptions of bolt damage and replacement work. The generator-bearing profiles preserve drive-end and non-drive-end locations and associate bearing problems with vibration and noise, although these indications alone do not establish rolling-contact fatigue.

InterBus hardware, converter-board identifiers, and fault codes retain source terminology (Table~\ref{tab:frequent_failure_modes}), while battery and accumulator profiles distinguish electrical and hydraulic energy storage (Table~\ref{tab:pitch_modes_comparison}). Candidate causes such as ageing, contamination, and thermal stress provide plausible engineering context.

The boundary between an intervention and a failure-related condition is apparent in \emph{Temporary oil cleanliness conditioning intervention}. Its candidate causes describe oil cleanliness assessment or contamination control, its common mechanisms describe temporary offline filtration, and its possible symptoms include filter installation, oil sampling, and subsequent removal. The profile identifies an activity without specifying a degraded item or an out-of-specification oil condition. A different boundary appears in \emph{Yaw motor overload alarm without confirmed hardware defect}: its possible symptoms include overload alarms and an inspection that found no abnormality. This profile preserves an indication without establishing a hardware defect.

Neighbouring categories can also overlap. \emph{Blade shell or coating damage requiring repair} includes local delamination within its common mechanisms (Table~\ref{tab:frequent_failure_modes}). \emph{Composite crack or laminate delamination} includes shell cracks, laminate damage, and structural repair actions. These descriptions do not provide a clear inclusion boundary between the profiles. The legacy Categories~3 and~4 descriptors in the former profile cannot resolve that boundary because their grading criteria are unavailable. Its candidate causes also include ``inspection-identified damage with exact cause not recorded''. This describes an observation and missing causal information rather than a candidate cause.

Internal coherence presents a further review question. \emph{Pitch brake assembly fails to release or hold as intended} combines opposite functional outcomes under one title (Table~\ref{tab:pitch_modes_comparison}). A broad label may be useful where the source does not distinguish those outcomes, but it does not preserve that distinction for later analysis. In the full profile for \emph{Pitch battery capacity loss or voltage collapse}, the possible symptoms include ``preventive full-string battery renewal'' alongside battery defect alarms, low autonomy, and voltage drop during battery tests. Renewal describes an intervention rather than a symptom.

Parent-system placement can reflect where a condition was observed. \emph{Hydraulic oil leakage into hub retention tray} is placed under \emph{Hub \& Main Shaft}, although its common mechanisms allow oil to originate in the pitch circuit (Table~\ref{tab:frequent_failure_modes}). The collection location therefore does not establish the originating component.

The selected cases identify category boundaries, field meanings, and system placements needing refinement before FMEA development. Section~\ref{subsec:reference_data_development} proposes source-based expert review. The reported labels and counts retain the original workflow outputs.

% ==============================================================================
% SECTION 4: DISCUSSION
% ==============================================================================
\section{Discussion}
\label{sec:discussion}

The workflow demonstrates fleet-scale semantic structuring with record-level provenance. Pitch separation, component-level actions, and distinctions between monitored equipment and monitoring hardware expose engineering workloads for review and future operational linkage. The balanced assessment estimates approval under explicit judgement rules and identifies changed maintenance types and failure-mode escapes as review priorities. Task yields reflect different input contexts and evidential demands; linked stages also permit early decisions to influence later labels and inclusion.

% ------------------------------------------------------------------------------
% SUBSECTION 4.1: LLM LIMITATIONS
% ------------------------------------------------------------------------------
\subsection{LLM limitations}
\label{subsec:llm_limitations}

The rejection synthesis (Section~\ref{subsubsec:qualitative_rejections}) and selected action and profile reviews (Section~\ref{subsec:semantic_label_review}) identify problems in both label application and category definition. Future iterations of the framework should prefer broader supported labels to unsupported diagnostic detail and separate conditions, mechanisms, symptoms, causes, and interventions. Independent reference judgements are needed to distinguish source ambiguity, taxonomy design, and model interpretation; revised prompts and taxonomies would require reassessment on separate records.

Approval estimates combine source-support judgements with unchanged-label retention. High approval of retained specific system codes and maintenance types indicates few clear contradictions, rather than the positive source confirmation required for changed labels.

The rates measure compliance with the assessment rules (Section~\ref{subsec:validation_design}), rather than accuracy against an independently adjudicated reference. Confidence intervals quantify sampling precision; assessor bias, interpretative uncertainty, and inter-expert agreement remain unmeasured. The qualitative synthesis elaborates the same judgements without estimating rejection-reason frequencies.

Expert reference labels are also interpretations of the available evidence. Hodkiewicz and Ho~\cite{hodkiewicz_cleaning_2016} document incomplete or unreliable functional-location and failure-mode fields in maintenance work orders. Walgern et al.~\cite{walgern_impact_2024} found that two organisations categorising the same wind-turbine reports obtained subsystem failure-rate ratios ranging from 0.71 to 3.5. Differences were associated with component-boundary interpretation, incomplete narratives, and annotator specialisation.

In text-annotation experiments, Wang et al.~\cite{wang_human-llm_2024} found that incorrect model suggestions could impair human judgement. These findings support documented coding rules and independent initial assessments. Consensus provides an agreed reference interpretation; confirmation of physical causes requires corroborating evidence.

Section~\ref{subsec:reference_data_development} proposes independent source-first labelling by multiple experts and documented resolution of disagreements (adjudication), alongside matched baseline comparisons.

Action and failure-mode taxonomies were derived from the archive they subsequently labelled, and the validation sample came from that archive. Distributions describe records within the proposed categories; taxonomy completeness and performance on unseen records require independent evaluation. Deterministic aggregation avoids batch-generated frequency estimates but remains dependent on category design.

Extraction instructions sought technically specific cases irrespective of frequency~\cite{malyi_2026_zenodo}. Retaining rare labels can increase taxonomy cardinality (the number of categories) and corroboration effort. Wording or inferred explanations can fragment related evidence, requiring source review and consolidation. These instructions may still leave unusual cases omitted or distinct phenomena combined.

Pre-processing standardises formatting and recurrent abbreviations; omitted observations, local expressions, and ambiguity still require source review. Text length alone does not determine usable evidence: short descriptions can support a label, while longer ones may add little relevant information.

Binary failure-mode assessment covered titles and escapes. Associated mechanisms, symptoms, and candidate causes remain profile-level context with variable record-level support. Title approval confirms neither every associated symptom nor a case-specific cause. Common denotes representative mechanisms, without establishing their prevalence.

The targeted pitch task relied on explicit pitch terms and operational acceptance. Poorly described pitch work may remain under legacy rotor or hydraulic codes. Overall system-code approval does not quantify this coverage gap or its effect on topology comparisons.

The controls in Section~\ref{subsubsec:acceptance_rules} govern routing and review: acceptance uses an uncalibrated confidence tier and a model-generated review flag, schemas check structure, and rationales provide stated model justifications. They establish neither semantic correctness nor database integrity. Independent source-first assessment should include routine unflagged outputs alongside referred cases (Section~\ref{subsec:reference_data_development}).

The low approval of failure-mode escapes (Section~\ref{subsec:validation_results}) shows that withholding a title or declaring inapplicability also requires scrutiny. Review should check both the evidence and the route to the status. Subgroup results pool the processing routes defined in Section~\ref{subsubsec:acceptance_rules} and cannot isolate their contributions. Descriptive system and action distributions are additionally conditional on candidate selection (Section~\ref{subsubsec:candidate_record_selection}).

Several records may concern one intervention, one record may describe several activities, and timestamps can mix engineering actions with administrative updates. Identifying independent physical failures, causes, and downtime therefore requires reconciliation with operational evidence (Section~\ref{subsec:future_research}).

Shared code, prompts, and schemas support reimplementation~\cite{malyi_2026_zenodo}; stored execution metadata and intermediate outputs support audit. Counts are traceable to record-level labels, but regenerating those labels depends on proprietary-model access, model version, prompts, taxonomies, reasoning settings, API behaviour, and exclusion rules. Model or API changes may alter outputs. Qualitative development checks did not quantify sensitivity to wording or reasoning settings.

System-code decisions govern taxonomy-extraction groups and later candidate dictionaries. Unsupported assignments can remove relevant profiles; a changed dictionary need not produce a different accepted label, since requests may return escapes. The separate influence of system context on maintenance-type inference was not measured. Joint rejections locate co-occurrence, not causal propagation. Without paired outputs under alternative upstream labels, the proportion of downstream changes attributable to system-code updates remains unknown. Section~\ref{subsec:reference_data_development} proposes controlled comparisons.

% ------------------------------------------------------------------------------
% SUBSECTION 4.2: IMPLEMENTATION TRADE-OFFS
% ------------------------------------------------------------------------------
\subsection{Implementation trade-offs}

% ..............................................................................
% SUBSUBSECTION 4.2.1: NLP APPROACHES AND WORKFLOW DESIGN
% ..............................................................................
\subsubsection{NLP approaches and workflow design}
\label{subsubsec:nlp_workflow_design}

Rule-based methods require explicit patterns or mappings, while supervised classifiers require representative labelled examples for training~\cite{salo_work_2019,lutz_kpi_2023}. Existing dictionaries, labelled datasets, or trained models may be reusable where their categories and source context match the intended task. Taxonomy detail and transfer between datasets remain assessment questions across approaches~\cite{walgern_impact_2024}. No task-specific model training was performed. Development involved prompt iteration, taxonomy preparation, schema design, and output assessment.

The shared code and schemas permit evaluation of alternative designs~\cite{malyi_2026_zenodo}. Joint inference could consider competing system assignments with their associated action and failure-mode candidates. Returning all fields together would still restrict selection if only one system's dictionary were supplied. Bounded iterative refinement could reconsider an earlier label after unsupported or inconsistent downstream results, retaining decision history and taxonomy versions. Both alternatives require evaluation of correctness, context demands, cost, and review effort, including cases where several fields share one unsupported interpretation.

Fine-tuned failure-mode coding~\cite{stewart_large_2023} and LLM-generated training examples for hierarchical classification~\cite{ding_large_2026} address defined classification tasks through different uses of labelled data. Taxonomy discovery additionally requires evaluation of the proposed categories. Fang et al.~\cite{fang_knowledge-guided_2024} used \texttt{GPT-4} and real-world evidence to construct a drug-indication taxonomy. Three experts assessed sampled relationships, finding higher accuracy for relationships between concepts than for mappings from extracted terms to concepts. These findings distinguish evaluation of the category structure from evaluation of individual mappings.

The present profiles similarly require review of both category relationships and individual mappings. Grouped evidence and individual accounts can support assessment of completeness, rare-case coverage, boundaries, and source support. Reviewed taxonomies could then serve manual labelling or bounded rule-based and supervised tools under the evaluation framework in Section~\ref{subsec:reference_data_development}.

% ..............................................................................
% SUBSUBSECTION 4.2.2: REVIEW ALLOCATION AND MEASURED WORKLOAD
% ..............................................................................
\subsubsection{Review allocation and measured workload}
\label{subsubsec:review_workload}

Deployment requires a review policy for ambiguous, unsupported, novel, or high-consequence outputs and independent samples of routine accepted labels. Measuring expert time would establish the effort allocated between structuring and assessment.

The workload measurements (Section~\ref{sec:project_economics}) characterise the selected model, execution-time prices, task structures, concurrency, and provider behaviour. Tokens describe billed text volume; internal computation, energy use, and infrastructure costs require separate measurements.

The reported token volumes provide a reference workload for approximate costing under alternative tariffs. For local deployment, they can support rough runtime estimates using representative input-processing and output-generation rates measured for the selected model and hardware under comparable context lengths and batching conditions~\cite{pope_efficiently_2022}. Estimates for another model should allow for differences in tokenisation, generated output length, and reasoning behaviour.

Beyond API charges, deployment costs include development, validation, expert review, infrastructure, governance, storage, security, and prompt and taxonomy maintenance. Comparison with manual or supervised alternatives requires human structuring and review times under equivalent quality criteria.

% ------------------------------------------------------------------------------
% SUBSECTION 4.3: TRANSFER CONDITIONS
% ------------------------------------------------------------------------------
\subsection{Transfer conditions}

Transfer beyond this wind-energy dataset requires local asset hierarchies, target taxonomies, configuration metadata, prompt revisions, privacy controls, and representative assessment records. Language, data-entry practices, failure definitions, and source-system structure may alter yield and validity. Even wind operators differ in RDS-PP extensions, work-order conventions, maintenance definitions, and topology metadata. The model-agnostic interface reduces provider-specific orchestration work; semantic transfer still requires local adaptation. Other industries using IBM Maximo offer settings for evaluating this transfer~\cite{ibm_maximo_2024}.

% ------------------------------------------------------------------------------
% SUBSECTION 4.4: MODEL AND REFERENCE-DATA DEVELOPMENT
% ------------------------------------------------------------------------------
\subsection{Model and reference-data development}
\label{subsec:reference_data_development}

A matched benchmark should compare rule- and dictionary-based methods, conventional supervised classifiers, fine-tuned transformers, and alternative LLMs. Compatible open-weight models, whose parameters are publicly available, could use the same task structure after interface adaptation. Compatibility requires sufficient context capacity, instruction following, and reliable structured responses.

Local deployment requires suitable computing resources: at a fixed numerical precision, models with more parameters need more memory for their weights, while throughput depends on hardware, context length, and batching~\cite{pope_efficiently_2022}. Smaller, potentially fine-tuned models may reduce resource demands, subject to task-specific assessment.

A multi-expert reference study should produce a reviewed taxonomy and independently adjudicated record-level labels. The panel should combine mechanical, electrical, hydraulic, and control expertise with knowledge of local practices, technicians' terminology, and documentation conditions.

Taxonomy review should assess system placement, topology, field meanings, and inclusion and exclusion criteria for neighbouring categories. Source evidence could justify retaining, splitting, merging, relabelling, or excluding profiles, preserving broad or unresolved categories where finer detail is unsupported. Decisions should retain rationales and taxonomy versions. Comparison with established domain taxonomies would assess shared concepts, granularity, and unmatched categories, separately from source support. Consolidation should preserve original profiles, source links, and mappings to the reviewed taxonomy.

Record-level reviewers should independently interpret source narratives before seeing model suggestions. Sampling should cover routine unflagged outputs, systems, taxonomy depths, sites, topologies, rare classes, language, text completeness, unlabelled records, and task routes. Field-level judgements, defensible alternatives, abstentions, and disagreement reasons should be retained before consensus adjudication, distinguishing system agreement from disagreement on component granularity or candidate cause.

The benchmark should use equivalent fields, definitions, source text, and equipment context, with adjudicated labels as the reference. Prompt refinement, rule construction, training, and taxonomy development should use development records only; taxonomies should be fixed before held-out evaluation. Discovery should be assessed separately from classification. Reporting should cover field-level correctness, rare categories, schema conformance, and abstention (declining to label). Labelled proportions and correctness among labelled records would distinguish coverage from validity. Profile descriptions require separate source-support assessment. Runtime, deployment conditions, and total development, processing, and expert-review costs should be assessed under common quality and referral criteria.

A propagation test should define a bounded cohort by asset and time independently of the labels, then obtain expert assessments of every included work order. Replacing upstream workflow labels with adjudicated labels would support two comparisons: inference with fixed dictionaries, and a broader rerun including affected taxonomy extraction. Retained taxonomy versions would distinguish category changes from record-level relabelling.

Both comparisons should rerun affected semantic stages, deterministic exclusions, and summaries while holding narratives, equipment metadata, and model configuration constant. Unchanged-input repeats would provide a reference for model variability. Comparing later fields, acceptance and escape statuses, eligibility, and label distributions would measure sensitivity. Source-based assessment would determine whether changes corrected or introduced unsupported interpretations.

Development records could support domain-specific fine-tuning and the use of retrieved reference material in prompts. Evaluation records should remain separate from these adaptations.

% ------------------------------------------------------------------------------
% SUBSECTION 4.5: FUTURE RESEARCH DIRECTIONS
% ------------------------------------------------------------------------------
\subsection{Future research directions}
\label{subsec:future_research}

Subsequent analyses should distinguish planned from performed activities and filter records according to the intended population, keeping records with unconfirmed completion separate. Operational analysis should reconstruct non-generation episodes from SCADA telemetry and alarm and status logs, associate eligible maintenance events, and attribute downtime under explicit temporal and equipment-compatibility rules. Separate accounting should distinguish events, unique associated episodes, and duration, counting each episode and its duration once per reporting group. Unmatched records and conflicts should remain visible for independent source-based review.

System and component labels can guide selection of alarm families and SCADA signals. Actions distinguish observation from restoration work, while documented symptoms suggest times for review. Disagreements with operational evidence would identify maintenance interpretations needing reassessment.

Topology reliability comparisons require reviewed source coverage, component installation and replacement histories, operating exposure, consistent event and episode definitions, uncertainty treatment, and consequence measures. FMEA would additionally require effects and existing controls. Occurrence requires an event unit and exposure denominator; severity requires a consequence measure, and detectability requires evidence about monitoring. The chosen method would determine normalisation, weighting, and combination for risk prioritisation.

A human-in-the-loop CMMS-entry interface could suggest codes and actions for review~\cite{wu_survey_2022}. It should distinguish planned from completed work, link related orders, and request components, observations, measured values, and units. Planning status, separate from maintenance type, plus permit or schedule references, campaign identifiers, and stoppage reasons would help distinguish campaigns from fault-led work. Timestamps should separate physical-work start and finish, operational restoration, and administrative closure; details could be marked unknown or inapplicable.

The interface should retain narratives, suggestions, supporting text, and reviewer acceptance, correction, or deferral, recording the reviewer, time, and taxonomy or model version. A prospective pilot should measure entry time, correction effort, missed or incorrect suggestions, validity, completeness, acceptance, and overrides. It should also assess governance, security, latency, and the consequences of incorrect suggestions.

% ==============================================================================
% SECTION 5: CONCLUSION
% ==============================================================================
\section{Conclusion}
\label{sec:conclusion}

This study implemented a topology-aware LLM workflow to review and enrich 16,316~maintenance records from 280~turbines across 31~onshore wind farms. Structured outputs, deterministic rules, and source links support traceable system, maintenance-type, action, and failure-mode labelling.

Accepted system-code updates covered 2,178 of 2,984~targeted records under the joint rule of \texttt{High} confidence and no human-review flag. Accepted failure-mode labels covered 11,661~records, comprising 10,654~turbine and 1,007~balance-of-plant records. API expenditure was \$368.86, and recorded task durations totalled 6.83~hours.

A source-based assessment of 800~records selected by balanced probability sampling yielded approval rates of 94.8\% for system codes, 87.3\% for maintenance types, 85.6\% for actions, and 77.5\% for failure-mode outcomes, covering titles and escape statuses.

Pitch separation and component-level labels support engineering case selection, while changed maintenance types, failure-mode escapes, action support, and profile definitions warrant further review. The contribution is an auditable semantic dataset for subsequent operational linkage, event reconstruction, reliability analysis, and FMEA development.

% ==============================================================================
% ADMIN SECTIONS AND BACK MATTER
% ==============================================================================

% ==============================================================================
% SECTION: DATA AVAILABILITY STATEMENT
% ==============================================================================
\section*{Data Availability Statement}
\label{sec:data_availability}

\noindent The Python workflow architecture, dynamic prompts, and data schemas are available for inspection and adaptation through \textcolor{blue}{\href{https://doi.org/10.5281/zenodo.20707310}{Zenodo}} and \textcolor{blue}{\href{https://github.com/mvmalyi/llm-driven-wind-turbine-maintenance-log-labelling}{GitHub}} under the MIT licence~\cite{malyi_2026_zenodo}. Commercial sensitivity prevents sharing the study data. Users of the framework are requested to cite this article.

% ==============================================================================
% SECTION: DATA GOVERNANCE AND EXTERNAL MODEL PROCESSING
% ==============================================================================
\section*{Data governance and external model processing}
\label{sec:data_governance}

\noindent The data provider permitted submission of the prepared maintenance records to external model APIs for this research. The original maintenance descriptors contained no personal information, including technician names. Before transmission, identifiers associated with the operator, wind farms, turbine locations, and original equipment manufacturers were removed or replaced with anonymised codes to protect commercial confidentiality. Manual review and automated Python checks before API submission identified no remaining sensitive information in the processed dataset. The 16,316-record corpus was processed through the OpenAI models using an API key supplied by EDINA at the University of Edinburgh and covered by a zero-data-retention arrangement.

% ==============================================================================
% SECTION: FUNDING
% ==============================================================================
\section*{Funding}

\noindent This study is part of an ongoing PhD project on wind turbine reliability, funded by the School of Engineering at the University of Edinburgh. The project received no other grants from any funding agency in the public, commercial, or not-for-profit sectors.

% ==============================================================================
% SECTION: ACKNOWLEDGEMENTS
% ==============================================================================
\section*{Acknowledgements}

\noindent The authors gratefully acknowledge EDINA, a centre for data and digital expertise at the University of Edinburgh, for supplying the OpenAI direct API key used in this study.

% ==============================================================================
% SECTION: DECLARATION OF COMPETING INTEREST
% ==============================================================================
\section*{Declaration of Competing Interest}

\noindent The authors declare no conflicts of interest.

% ==============================================================================
% SECTION: DECLARATION OF GENERATIVE AI AND AI-ASSISTED TECHNOLOGIES IN THE MANUSCRIPT PREPARATION PROCESS
% ==============================================================================
\section*{Declaration of generative AI \\ and AI-assisted technologies in the manuscript preparation process}

\noindent During the preparation of this work, the authors used Google \texttt{gemini-3.1-pro} to assist with language editing, stylistic refinement, and proofreading of the manuscript text. OpenAI \texttt{gpt-5.4-thinking} supported the optimisation and debugging of the accompanying Python scripts. After using these tools, the authors reviewed and edited the content as needed and take full responsibility for the content of the published article.

% ==============================================================================
% SECTION: CREDIT AUTHORSHIP CONTRIBUTION STATEMENT
% ==============================================================================
\section*{CRediT Authorship Contribution Statement}

\noindent \textbf{Max Malyi:} Conceptualisation, Formal analysis, Investigation, Methodology, Software, Visualisation, Validation, Writing–original draft, Writing–review \& editing. \textbf{Jonathan Shek:} Resources, Supervision, Writing–review \& editing. \textbf{Alasdair McDonald:} Supervision, Writing–review \& editing. \textbf{André Biscaya:} Data curation, Supervision, Writing–review \& editing.

% ==============================================================================
% DEFINITIONS
% ==============================================================================

% ==============================================================================
% REFERENCES
% ==============================================================================
\phantomsection
\addcontentsline{toc}{section}{References}
\bibliographystyle{references-style}
\bibliography{references}

@article{alhmoud_review_2018,
    author   = {Alhmoud, Lina and Wang, Bingsen},
    title    = {A Review of the State-of-the-Art in Wind-Energy Reliability Analysis},
    journal  = {Renewable and Sustainable Energy Reviews},
    year     = {2018},
    month    = jan,
    volume   = {81},
    pages    = {1643--1651},
    doi      = {10.1016/j.rser.2017.05.252},
    urldate  = {2023-07-22},
}

@article{artigao_failure_2021,
    author   = {Artigao, Estefania and Martin-Martinez, Sergio and Ceña, Alberto and Honrubia-Escribano, Andres and Gomez-Lazaro, Emilio},
    title    = {Failure Rate and Downtime Survey of Wind Turbines Located in {Spain}},
    journal  = {{IET} Renewable Power Generation},
    year     = {2021},
    month    = jan,
    volume   = {15},
    number   = {1},
    pages    = {225--236},
    doi      = {10.1049/rpg2.12019},
    urldate  = {2025-01-06},
}

@article{carroll_failure_2016,
    author   = {Carroll, James and McDonald, Alasdair and McMillan, David},
    title    = {Failure Rate, Repair Time and Unscheduled {O\&M} Cost Analysis of Offshore Wind Turbines},
    journal  = {Wind Energy},
    year     = {2016},
    month    = jun,
    volume   = {19},
    number   = {6},
    pages    = {1107--1119},
    doi      = {10.1002/we.1887},
    urldate  = {2023-09-19},
}

@article{deville_cube_2004,
    author   = {Deville, Jean-Claude and Till\'{e}, Yves},
    title    = {Efficient Balanced Sampling: The Cube Method},
    journal  = {Biometrika},
    year     = {2004},
    month    = dec,
    volume   = {91},
    number   = {4},
    pages    = {893--912},
    doi      = {10.1093/biomet/91.4.893},
}

@article{deville_variance_2005,
    author   = {Deville, Jean-Claude and Till\'{e}, Yves},
    title    = {Variance Approximation under Balanced Sampling},
    journal  = {Journal of Statistical Planning and Inference},
    year     = {2005},
    month    = feb,
    volume   = {128},
    number   = {2},
    pages    = {569--591},
    doi      = {10.1016/j.jspi.2003.11.011},
}

@article{dinmohammadi_fuzzy-fmea_2013,
    author   = {Dinmohammadi, F. and Shafiee, M.},
    title    = {A Fuzzy-{FMEA} Risk Assessment Approach for Offshore Wind Turbines},
    journal  = {International Journal of Prognostics and Health Management},
    year     = {2013},
    volume   = {4},
    number   = {3},
    pages    = {1--10},
    doi      = {10.36001/ijphm.2013.v4i3.2143},
}

@article{leahy_issues_2019,
    author   = {Leahy, Kevin and Gallagher, Colm and O'Donovan, Peter and O'Sullivan, Dominic T. J.},
    title    = {Issues with Data Quality for Wind Turbine Condition Monitoring and Reliability Analyses},
    journal  = {Energies},
    year     = {2019},
    month    = jan,
    volume   = {12},
    number   = {2},
    eid      = {201},
    doi      = {10.3390/en12020201},
    urldate  = {2023-07-22},
}

@article{li_improved_2022,
    author   = {Li, He and Teixeira, A. P. and Guedes Soares, C.},
    title    = {An Improved Failure Mode and Effect Analysis of Floating Offshore Wind Turbines},
    journal  = {Journal of Marine Science and Engineering},
    year     = {2022},
    month    = nov,
    volume   = {10},
    number   = {11},
    eid      = {1616},
    doi      = {10.3390/jmse10111616},
    urldate  = {2023-10-24},
}

@article{lutz_digitalization_2022,
    author  = {Lutz, Marc-Alexander and Walgern, Julia and Beckh, Katharina and Schneider, Juliane and Faulstich, Stefan and Pfaffel, Sebastian},
    title   = {Digitalization Workflow for Automated Structuring and Standardization of Maintenance Information of Wind Turbines into Domain Standard as a Basis for Reliability {KPI} Calculation},
    journal = {Journal of Physics: Conference Series},
    year    = {2022},
    month   = apr,
    volume  = {2257},
    number  = {1},
    eid     = {012004},
    doi     = {10.1088/1742-6596/2257/1/012004},
    urldate = {2025-03-14},
}

@article{lutz_kpi_2023,
    author   = {Lutz, Marc-Alexander and Schäfermeier, Bastian and Sexton, Rachael and Sharp, Michael and Dima, Alden and Faulstich, Stefan and Aluri, Jagan Mohini},
    title    = {{KPI} Extraction from Maintenance Work Orders: A Comparison of Expert Labeling, Text Classification and {AI}-Assisted Tagging for Computing Failure Rates of Wind Turbines},
    journal  = {Energies},
    year     = {2023},
    month    = dec,
    volume   = {16},
    number   = {24},
    eid      = {7937},
    doi      = {10.3390/en16247937},
    urldate  = {2025-07-08},
}

@inproceedings{salo_work_2019,
    author    = {Salo, Erik and McMillan, David and Connor, Richard},
    title     = {Work Orders: Value from Structureless Text in the Era of Digitisation},
    booktitle = {{SPE} Offshore Europe Conference and Exhibition},
    year      = {2019},
    month    = sep,
    paper     = {SPE-195788-MS},
    publisher = {{SPE}},
    address   = {Aberdeen, UK},
    doi       = {10.2118/195788-MS},
    urldate   = {2025-03-14},
}

@article{scheu_systematic_2019,
    author   = {Scheu, Matti Niclas and Tremps, Lorena and Smolka, Ursula and Kolios, Athanasios and Brennan, Feargal},
    title    = {A Systematic Failure Mode Effects and Criticality Analysis for Offshore Wind Turbine Systems Towards Integrated Condition Based Maintenance Strategies},
    journal  = {Ocean Engineering},
    year     = {2019},
    month    = mar,
    volume   = {176},
    pages    = {118--133},
    doi      = {10.1016/j.oceaneng.2019.02.048},
    urldate  = {2023-07-22},
}

@article{tazi_using_2017,
    author   = {Tazi, Nacef and Châtelet, Eric and Bouzidi, Youcef},
    title    = {Using a Hybrid Cost-{FMEA} Analysis for Wind Turbine Reliability Analysis},
    journal  = {Energies},
    year     = {2017},
    month    = feb,
    volume   = {10},
    number   = {3},
    eid      = {276},
    doi      = {10.3390/en10030276},
    urldate  = {2023-10-24},
}

@article{walgern_impact_2024,
    author   = {Walgern, Julia and Beckh, Katharina and Hannes, Neele and Horn, Martin and Lutz, Marc-Alexander and Fischer, Katharina and Kolios, Athanasios},
    title    = {Impact of Using Text Classifiers for Standardising Maintenance Data of Wind Turbines on Reliability Calculations},
    journal  = {{IET} Renewable Power Generation},
    year     = {2024},
    month    = nov,
    volume   = {18},
    number   = {15},
    pages    = {3463--3479},
    doi      = {10.1049/rpg2.13151},
    urldate  = {2025-03-14},
}

@article{walker_using_2024,
    author  = {Walker, C. and Rothon, C. and Aslansefat, K. and Papadopoulos, Y. and Dethlefs, N.},
    title   = {Using Large Language Models to Recommend Repair Actions for Offshore Wind Maintenance},
    journal = {Journal of Physics: Conference Series},
    year    = {2024},
    month   = nov,
    volume  = {2875},
    number  = {1},
    eid     = {012025},
    doi     = {10.1088/1742-6596/2875/1/012025},
    urldate = {2025-07-08},
}

@misc{walshe_automatic_2025,
    author       = {Walshe, Thomas and Moon, Sae Young and Xiao, Chunyang and Gunawardana, Yawwani and Silavong, Fran},
    title        = {Automatic Labelling with Open-Source {LLMs} Using Dynamic Label Schema Integration},
    howpublished = {arXiv preprint arXiv:2501.12332 [cs]},
    year         = {2025},
    month    = jan,
    doi          = {10.48550/arXiv.2501.12332},
}

@article{wang_integrated_2022,
    author   = {Wang, Zhen and Wang, Rongxi and Deng, Wei and Zhao, Yong},
    title    = {An Integrated Approach-Based {FMECA} for Risk Assessment: Application to Offshore Wind Turbine Pitch System},
    journal  = {Energies},
    year     = {2022},
    month    = mar,
    volume   = {15},
    number   = {5},
    eid      = {1858},
    doi      = {10.3390/en15051858},
    urldate  = {2025-11-28},
}

@inproceedings{wilkinson_european_2010,
    author    = {Wilkinson, Michael and Hendriks, Ben and Spinato, Fabio and Gomez, Eugenio and Bulacio, Horacio and Roca, Jordi and Tavner, Peter and Feng, Yanhui and Long, Hui},
    title     = {Methodology and Results of the {Reliawind} Reliability Field Study},
    booktitle = {European Wind Energy Conference},
    year      = {2010},
    pages     = {1984--1990},
    address   = {Warsaw, Poland},
    url       = {https://eprints.whiterose.ac.uk/id/eprint/83343/},
}

@standard{VGB_RDSPP_Part32_2021,
    author      = {{VGB PowerTech e.V.}},
    title       = {{RDS-PP\textregistered{}} Application Guideline: Part 32: Wind Power Plants},
    number      = {VGB-S-823-32-2021-12-EN-DE},
    edition     = {2},
    year        = {2021},
    institution = {{VGB PowerTech e.V.}},
    publisher   = {{VGB PowerTech Service GmbH}},
    address     = {Essen, Germany},
    url         = {https://shop.vgbe.energy/s-823-32.html},
}

@article{malyi_exploratory_2026,
    author   = {Malyi, Max and Shek, Jonathan and Biscaya, André},
    title    = {Exploratory Semantic Reliability Analysis of Wind Turbine Maintenance Logs Using Large Language Models},
    journal  = {{IET} Renewable Power Generation},
    year     = {2026},
    volume   = {20},
    number   = {1},
    eid      = {e70327},
    doi      = {10.1049/rpg2.70327},
}

@article{malyi_comparative_2026,
    author   = {Malyi, Max and Shek, Jonathan and McDonald, Alasdair and Biscaya, André},
    title    = {A Comparative Benchmark of Large Language Models for Labelling Wind Turbine Maintenance Logs},
    journal  = {{IET} Renewable Power Generation},
    year     = {2026},
    volume   = {20},
    number   = {1},
    eid      = {e70212},
    doi      = {10.1049/rpg2.70212},
}

@techreport{openai_gpt54_2026,
    author      = {{OpenAI}},
    title       = {{GPT-5.4} Thinking System Card},
    institution = {{OpenAI}},
    year        = {2026},
    month    = mar,
    url         = {https://deploymentsafety.openai.com/gpt-5-4-thinking/gpt-5-4-thinking.pdf},
    note        = {Accessed 9 May 2026},
}

@misc{openai_python_2025,
    author   = {{OpenAI}},
    title    = {{OpenAI} {Python} {API} Library},
    year     = {2025},
    url      = {https://developers.openai.com/api/reference/python},
    note     = {Accessed 9 May 2026},
}

@misc{ibm_maximo_2024,
    author = {{IBM}},
    title  = {{IBM} Maximo Application Suite},
    year   = {2024},
    url    = {https://www.ibm.com/products/maximo},
    note   = {Accessed 9 May 2026},
}

@article{lin_fault_2016,
    author  = {Lin, Yonggang and Tu, Le and Liu, Hongwei and Li, Wei},
    title   = {Fault Analysis of Wind Turbines in {China}},
    journal = {Renewable and Sustainable Energy Reviews},
    year    = {2016},
    month   = mar,
    volume  = {55},
    pages   = {482--490},
    doi     = {10.1016/j.rser.2015.10.149},
}

@misc{malyi_2026_zenodo,
    author       = {Malyi, Max and Shek, Jonathan and McDonald, Alasdair and Biscaya, André},
    title        = {Supporting Open-Source Repository for Wind Turbine Maintenance Log Labelling Framework: {LLM}-Driven Data Correction and Enrichment via Semantic Extraction of Reliability Intelligence},
    howpublished = {Zenodo},
    year         = {2026},
    doi          = {10.5281/zenodo.20707310},
    url          = {https://doi.org/10.5281/zenodo.20707310},
}

@inproceedings{malyi_wind_2025,
    author    = {Malyi, Max and Shek, Jonathan and McDonald, Alasdair},
    title     = {Wind Turbine Reliability Analysis Using {SCADA} and Maintenance Data},
    booktitle = {Solar and Wind Beyond Limits for Technology, Policy, and Practice},
    editor    = {Muhammad-Sukki, Firdaus and Sellami, Nazmi},
    year      = {2025},
    pages     = {185--193},
    publisher = {Springer Nature},
    address   = {Edinburgh, UK},
    doi       = {10.1007/978-3-032-08953-3_18},
}

@article{wu_survey_2022,
    author  = {Wu, Xingjiao and Xiao, Luwei and Sun, Yixuan and Zhang, Junhang and Ma, Tianlong and He, Liang},
    title   = {A Survey of Human-in-the-Loop for Machine Learning},
    journal = {Future Generation Computer Systems},
    year    = {2022},
    month   = oct,
    volume  = {135},
    pages   = {364--381},
    doi     = {10.1016/j.future.2022.05.014},
}

@misc{stewart_large_2023,
    author       = {Stewart, Michael and Hodkiewicz, Melinda and Li, Sirui},
    title        = {Large Language Models for Failure Mode Classification: An Investigation},
    howpublished = {arXiv preprint arXiv:2309.08181 [cs.CL]},
    year         = {2023},
    month        = sep,
    doi          = {10.48550/arXiv.2309.08181},
    urldate      = {2026-10-20},
}

@article{ding_large_2026,
    author   = {Ding, Yi and Zhu, Feng and Li, He and Xie, Min},
    title    = {A Large Language Model-Augmented Hierarchical Text Classification Approach for Intelligent Failure Data Management},
    journal  = {Engineering Applications of Artificial Intelligence},
    year     = {2026},
    month    = oct,
    volume   = {181},
    eid      = {115750},
    doi      = {10.1016/j.engappai.2026.115750},
    urldate  = {2026-10-20},
}

@article{men_interpretable_2025,
    author   = {Men, ChangHao and Han, Yu and Wang, Ping and Tao, JianYuan and Huang, Cheng-Geng},
    title    = {The Interpretable Reasoning and Intelligent Decision-Making Based on Event Knowledge Graph with {LLMs} in Fault Diagnosis Scenarios},
    journal  = {{IEEE} Transactions on Instrumentation and Measurement},
    year     = {2025},
    volume   = {74},
    pages    = {1--16},
    doi      = {10.1109/TIM.2025.3550999},
    urldate  = {2026-09-11},
}

@standard{en_maintenance_terminology_2017,
    author      = {{European Committee for Standardization}},
    title       = {Maintenance: Maintenance Terminology},
    number      = {EN 13306:2017},
    year        = {2017},
    month       = nov,
    institution = {{European Committee for Standardization}},
    address     = {Brussels, Belgium},
    url         = {https://landingpage.bsigroup.com/LandingPage/Standard?UPI=000000000030324472},
    urldate     = {2026-10-17},
}

@article{walgern_reliability_2023,
    author   = {Walgern, Julia and Fischer, Katharina and Hentschel, Paul and Kolios, Athanasios},
    title    = {Reliability of Electrical and Hydraulic Pitch Systems in Wind Turbines Based on Field-Data Analysis},
    journal  = {Energy Reports},
    year     = {2023},
    month    = dec,
    volume   = {9},
    pages    = {3273--3281},
    doi      = {10.1016/j.egyr.2023.02.007},
    urldate  = {2023-07-22},
}

@misc{pope_efficiently_2022,
    author       = {Pope, Reiner and Douglas, Sholto and Chowdhery, Aakanksha and Devlin, Jacob and Bradbury, James and Levskaya, Anselm and Heek, Jonathan and Xiao, Kefan and Agrawal, Shivani and Dean, Jeff},
    title        = {Efficiently Scaling {Transformer} Inference},
    howpublished = {arXiv preprint arXiv:2211.05102 [cs.LG]},
    year         = {2022},
    month        = nov,
    doi          = {10.48550/arXiv.2211.05102},
    urldate      = {2026-10-27},
}

@article{hodkiewicz_cleaning_2016,
    author   = {Hodkiewicz, Melinda and Ho, Mark Tien-Wei},
    title    = {Cleaning Historical Maintenance Work Order Data for Reliability Analysis},
    journal  = {Journal of Quality in Maintenance Engineering},
    year     = {2016},
    month    = may,
    volume   = {22},
    number   = {2},
    pages    = {146--163},
    doi      = {10.1108/JQME-04-2015-0013},
    urldate  = {2025-07-08},
}

@inproceedings{wang_human-llm_2024,
    author    = {Wang, Xinru and Kim, Hannah and Rahman, Sajjadur and Mitra, Kushan and Miao, Zhengjie},
    title     = {Human-{LLM} Collaborative Annotation Through Effective Verification of {LLM} Labels},
    booktitle = {Proceedings of the {CHI} Conference on Human Factors in Computing Systems},
    year      = {2024},
    month     = may,
    pages     = {1--21},
    publisher = {ACM},
    address   = {Honolulu, HI, USA},
    doi       = {10.1145/3613904.3641960},
}

@article{fang_knowledge-guided_2024,
    author   = {Fang, Yilu and Ryan, Patrick and Weng, Chunhua},
    title    = {Knowledge-Guided Generative Artificial Intelligence for Automated Taxonomy Learning from Drug Labels},
    journal  = {Journal of the American Medical Informatics Association},
    year     = {2024},
    month    = sep,
    volume   = {31},
    number   = {9},
    pages    = {2065--2075},
    doi      = {10.1093/jamia/ocae105},
}

@article{youn_linear_2025,
    author   = {Youn, Yeochan and Schulz, Joseph B. and Stolen, Ethan and Farrey, Karl and Muresan, Emil and Kim, Siyong and Sohn, James J.},
    title    = {Linear Accelerator ({Linac}) Downtime Analysis Assisted with a Large Language Model ({LLM})},
    journal  = {Journal of Applied Clinical Medical Physics},
    year     = {2025},
    month    = dec,
    volume   = {26},
    number   = {12},
    eid      = {e70400},
    doi      = {10.1002/acm2.70400},
    urldate  = {2026-10-20},
}

\end{document}